\documentclass[12pt,3p]{elsarticle}
\usepackage[utf8]{inputenc}

\usepackage[namelimits,sumlimits]{amsmath}
\usepackage{amsfonts,amssymb,amsthm}
\usepackage{cancel} % For strike through arguments without affecting the bibliography style, don't replace with \usepackage{ulem}
\usepackage{soul} % Warping the underline text, use \ul{} instead of \underline
\usepackage{graphicx}
\graphicspath{{images/}}
\usepackage[hidelinks]{hyperref}
\usepackage{fancyhdr}
\usepackage{url}
\usepackage{array}
\usepackage{dirtree}
\usepackage{enumitem}
\usepackage[explicit]{titlesec}
\usepackage{etoolbox}
\usepackage{acro}
\usepackage{xcolor}
\usepackage{subcaption}
\newsavebox{\imagebox}
\usepackage{blindtext}
\usepackage{csquotes}
\usepackage{pgfgantt}
\usepackage[export]{adjustbox}
\usepackage{ragged2e}
\usepackage{pdflscape} % provides the landscape environment
\usepackage{tikz}
\usetikzlibrary{shapes.geometric, arrows}
\usetikzlibrary{arrows.meta}
\usetikzlibrary{quotes,angles}
\usepackage{textpos}
\usepackage[linesnumbered,ruled,vlined]{algorithm2e}
\usepackage{float}
\usepackage{import}
\usepackage{amsthm}

\newcommand{\captioncomment}[2]{\caption[{#1}]{#1\\ #2 }}

\newcommand{\changes}[1]{\textcolor{black}{#1}}
\newtheorem{thm}{Theorem}

\newtheorem{propos}{Proposition}
\newtheorem{lemma}{Lemma}

\newtheorem{defin}{Definition}

\date{June 2020}

\begin{document}
\title{Necessary and sufficient condition for a generic 3R serial manipulator to be cuspidal}
\author[add1]{Durgesh Haribhau Salunkhe}
    \ead{durgesh.salunkhe@ls2n.fr}
\author[add2]{Christoforos Spartalis}
  \ead{christoforos.spartalis@uibk.ac.at}
  \author[add2]{Jose Capco}
  \ead{jose.capco@uibk.ac.at}
  \author[add1]{Damien Chablat}
  \ead{damien.chablat@cnrs.fr}
  \author[add1]{Philippe Wenger\corref{cor1}}
  \ead{philippe.wenger@ls2n.fr}

  \cortext[cor1]{Please address correspondence to Philippe Wenger}
  \address[add1]{REV team, Laboratoire des Sciences du Num\'erique de Nantes, France}
  \address[add2]{University of Innsbruck, Austria}
\begin{abstract}
    Cuspidal robots can travel from one inverse kinematic solution to another without meeting a singularity. The name cuspidal was coined based on the existence of a cusp point in the workspace of 3R serial robots. The existence of a cusp point was proved to be a necessary and sufficient condition for \emph{orthogonal} robots to be cuspidal, but it was not possible to extend this condition to non-orthogonal robots. The goal of this paper is to prove that this condition stands for \emph{any} generic 3R robot. This result would give the designer more flexibility. In the presented work, the geometrical interpretation of the inverse kinematics of 3R robots is revisited and important observations on the nonsingular change of posture are noted.  The paper presents a theorem regarding the existence of reduced aspects in any generic 3R serial robot. Based on these observations and on this theorem, we prove that the existence of a cusp point is a necessary and sufficient condition for any 3R generic robot to be cuspidal.
\end{abstract}

\maketitle
 %%%%%%%%%%% section: introduction %%%%%%%%%%%
 \section{Introduction}
\label{section:introduction}  
Cuspidal robots are those robots that can travel from one inverse kinematic solution (IKS) to another without encountering a singularity. This property of cuspidal robots is referred to as cuspidality. The name `\textit{cuspidal}' originated from the existence of a cusp in the singularity locus in the workspace of the robot \cite{el1995recognize}. Cuspidality exists in both serial and parallel robots \cite{innocenti1998singularity} but this paper focuses on serial robots only. Cuspidality was first identified in 1988 by Parenti-Castelli \cite{parenti-castelli_position_1988} in some 6R robots and by Burdick in 1989 \cite{burdick_inverse_1989} in 3R robots. Most industrial robots are noncuspidal and the posture in which a noncuspidal robot is operating can be easily identified with the signs of the factors of the determinant of the Jacobian matrix \cite{borrel_study_1986}.  Instead, posture identification is very difficult in cuspidal robots \changes{as the determinant does not usually factor}  \cite{wenger_new_1992}.This makes trajectory planning more challenging \cite{wenger_uniqueness_2004}. Cuspidal robots were first formalized and later extensively studied by Wenger et al. \cite{wenger_new_1992, wenger_comments_1997, baili_classification_2004, wenger_uniqueness_2004, wenger_dh-parameter_2005, wenger_cuspidal_2007, wenger_cuspidal_2019}. Different approaches were implemented in the past to identify and classify 3R \emph{orthogonal} robots (i.e. robots with three mutually orthogonal joint axes), based on cuspidality. 

\changes{Identifying the number of aspects can allow the designer to identify whether a given robot is cuspidal or not.} Paganelli \cite{paganelli_topological_2008} and Wenger et al. \cite{wenger:hal-02362907} proposed a homotopy based topological analysis of singularity loci to identify the maximum number of aspects for a regional 3R serial chain. Though useful in many cases, this approach cannot be implemented to classify robots based on cuspidality since the number of cusps is not constant in a given homotopy class \cite{baili_analyse_2004}. Baili proposed a deeper analysis and exhaustive classification of 3R positional  \emph{orthogonal} serial robots based on cuspidality. This class of robots can be analyzed in detail by using different algebraic and topological tools. This is because the orthogonality constraint helps simplify some coefficients in the inverse kinematics polynomial \cite{kohli_workspace_1985}.  

\begin{figure}
	\centering
	\includegraphics[width = 0.7\textwidth]{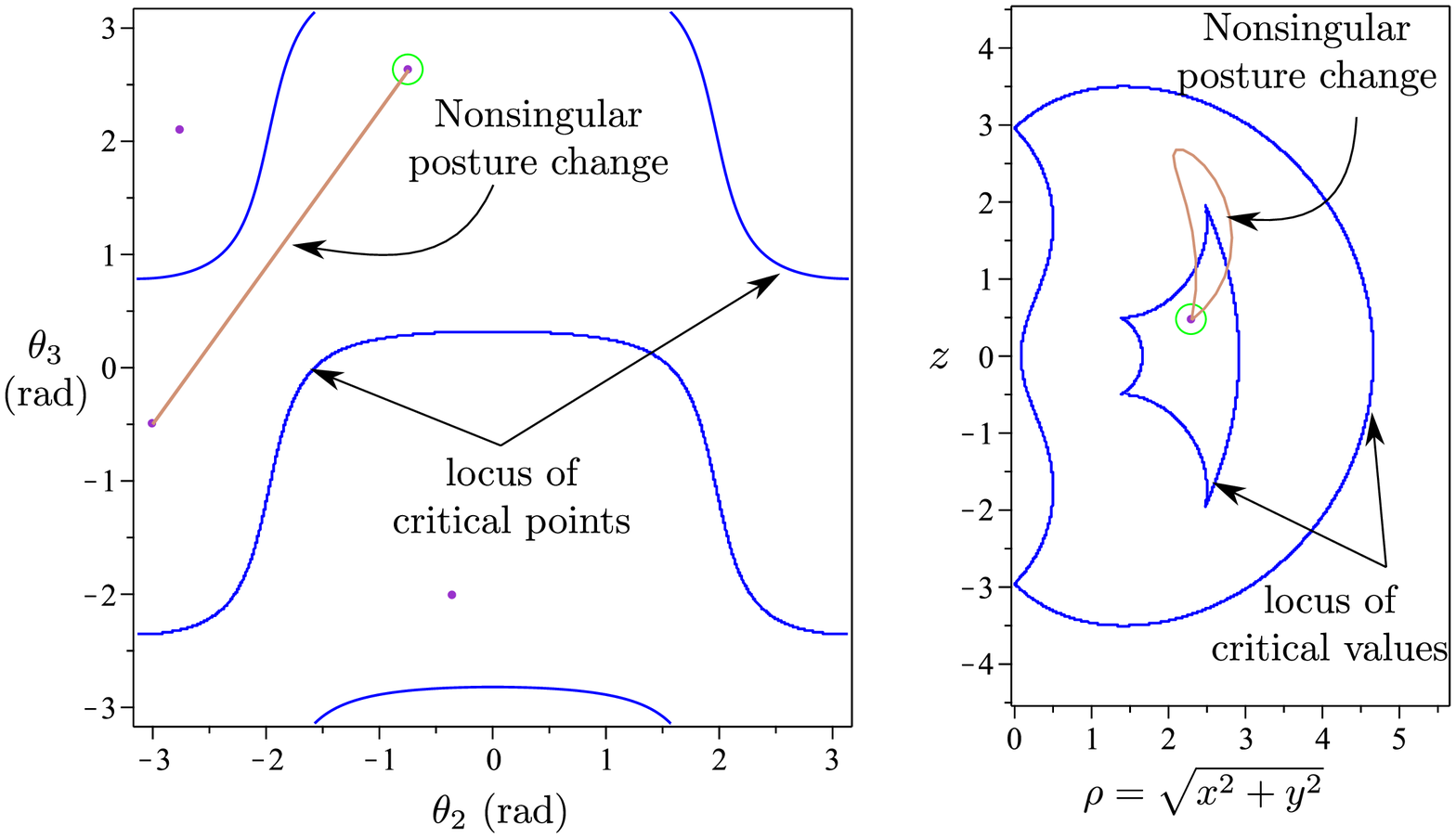}
	\caption{An example of travelling from one IKS to another in joint space and workspace.\\Robot parameters: d = [0, 1, 0], a = [1, 2, 3/2], $\alpha$ = [$-\dfrac{\pi}{2}$, $\dfrac{\pi}{2}$, 0].\\path in the joint space ($\theta_2$, $\theta_3$): from ($-0.742, 2.628$) to ($-3, -0.5$).}
	\label{fig:nonsingular_traj}
\end{figure}

The presence of a cusp in the workspace was proved to be a necessary and sufficient condition for cuspidality in 3R \emph{orthogonal} robots \cite{wenger_cuspidal_2019}. The proof used the fact that the complete parameter space of orthogonal robots was mapped, and it was clear from the classification that the nonsingular change of IKS meant \changes{encircling a cusp} in the workspace \cite{wenger_changing_1996} as illustrated in Fig. \ref{fig:nonsingular_traj}.
Algebraic analysis of non-orthogonal 3R robots is more challenging, and no classification scheme has been attempted yet. In the absence of any counter-example, it has been conjectured that non-orthogonal 3R robots should behave like their orthogonal counterparts, i.e., they should have a cusp in the workspace to be cuspidal, but no formal proof exists to confirm this conjecture. A cusp allows a \emph{local} nonsingular change of posture in any 3R robot \cite{corvez_study_2005} and this shows that the existence of a cusp is a \emph{sufficient} condition for any 3R robot to be cuspidal. However, in theory, a nonsingular change of solution could be also feasible in a more global way and without any cusp. In fact, this feature was shown to be present in parallel robots \cite{coste2014nonsingular}. 

A necessary and sufficient condition \changes{of cuspidality for \emph{any generic} 3R robot to be cuspidal} would allow great deal of flexibility to the designers to choose from parameters that do not require strict alignment, orthogonality or intersection of joint axes. 

The aim of this work is to provide a formal proof that the existence of a cusp \changes{is a necessary and sufficient condition} for \emph{any generic} 3R robot to be cuspidal, should they be orthogonal or not.
%Results obtained from such a classification could be used for 6R wrist-partitioned robots, which are most of the implemented robots in the industry. Those robots, moreover, have either parallel intersecting joint axes. These arrangements simplify the kinematic analysis but put forth several challenges in manufacturing and demand strict tolerances. 
This proof can be extended to all classes of 6R robots where the \changes{position} degrees of freedom (dof) are decoupled from the orientation dof, such as 6R robots with a spherical wrist. This class forms a \changes{large} population of 6R robots, thus emphasizing the impact of the presented work.

The following work is divided into three sections: Section \ref{section:preliminaries} revisits the geometrical interpretation of the inverse kinematics of 3R robots, singularities and nonsingular change of posture. These concepts are then presented both in joint space and workspace in order to draw parallels. The main contribution of the work is expounded in Section \ref{section:proof} where a necessary and sufficient cuspidality condition for any generic 3R robot is put forth. Section \ref{section:conclusions} concludes the work by discussing the implications of the contribution and addressing a few pointers to future work.

 %%%%%%%%%%% section: implementation %%%%%%%%%%%
 \section{Preliminaries}
\label{section:preliminaries}  
This section revisits briefly the geometric interpretation of inverse kinematic solutions proposed by Pieper \cite{pieper_kinematics_1968}. This geometric interpretation will be used for our proof. Then, the interpretation of critical points (singularities) and nonsingular change of posture in the joint space and workspace are discussed, along with their geometrical implications. Relevant definitions and their interpretations in different spaces are explained in order to provide a background to the proof of the necessary and sufficient cuspidality condition. The section also highlights key terms relevant to the proposed proof.
\subsection{Inverse kinematic solutions}
 Let $\mathbf{x}=(x,y,z)$ be the vector of coordinates of the robot's end effector in the workspace  $\mathcal{W} \subset \mathbb{R}^3$ at a configuration $\mathbf{q}=(\theta_1,\theta_2,\theta_3)$ in the joint space $\mathcal{J}=\mathcal{S}^1 \times \mathcal{S}^1 \times \mathcal{S}^1$.  The mapping between $\mathcal{J}$ and $\mathcal{W}$, denoted by $f:  \mathcal{J} \rightarrow \mathcal{W}$, defines the direct kinematics \eqref{eq:dkf}. 
\begin{equation}
    \mathbf{x} = f(\mathbf{q}), \mathbf{x} \in \mathcal{W}, \mathbf{q} \in \mathcal{J}
    \label{eq:dkf}
\end{equation}

The elements in the preimage $f^{-1}(\mathbf{q})$ are the inverse kinematic solutions (IKS) of $\mathbf{q}$. A robot configuration associated with an IKS is called a \emph{posture}.

\begin{figure}
    \centering
    \includegraphics[width = 0.5 \textwidth]{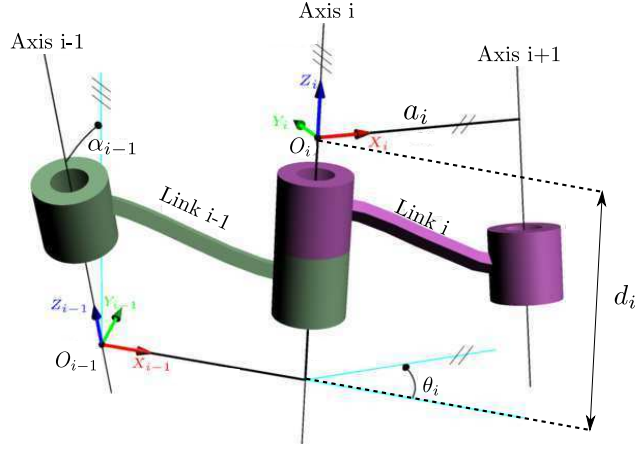}
    \caption{The D-H parameter notations used.}
    \label{fig:dhpara}
\end{figure}

 Solving the inverse kinematics of 3R serial robots was first reported in \cite{pieper_kinematics_1968} where it was noted that the solutions correspond to the intersection of a conic with a circle in $c_3s_3$-plane, where $c_3$ and $s_3$ denote $\cos\theta_3$ and $\sin\theta_3$, respectively. The solution is presented briefly, as it has a key role in the proof to follow. 
In this paper, classical D-H parameters are used, as shown in Fig. \ref{fig:dhpara}.

 Let, $R = \rho^2 + z^2,$ where $\rho^2 = x^2 + y^2= g(\theta_2, \theta_3).$  
The terms $R$ and $z$ can be written as 
\begin{align*}
    R &= (F_1\,\cos\theta_2 + F_2\,\sin\theta_2)\,2a_1 + F_3 \\
    z &= (F_1\,\sin\theta_2 - F_2\,\cos\theta_2)\,\sin\alpha_1 + F_4
\end{align*}where $F_i = g_i(\theta_3),$ for $i=1,..,4.$ Upon rearrangement, we obtain the general equation of a conic in $c_3s_3$-plane as given in (\ref{eq:conic_gen}).
\begin{equation}
A_{xx}\,c_3^2 + 2A_{xy}\,c_3s_3 + A_{yy}\,s_3^2 + 2B_x\,c_3 + 2B_y\,s_3 + C = 0
\label{eq:conic_gen}
\end{equation}
The coefficients of the conic are skipped for brevity, but they are functions of the D-H parameters and of $(R,z)$ as shown in (\ref{eq:conic_coeffs}),
\begin{equation}
	\begin{aligned}
		A_{xx} &= h_1(a_1, a_2, a_3) \\
		A_{xy} &= h_2(a_1, a_2, a_3, d_2, \alpha_2)\\
		A_{yy} &= h_3(a_1, a_2, a_3, d_2, \alpha_1, \alpha_2)\\
		B_{x} &= h_4(a_1, a_2, a_3, d_2, \alpha_2, R)\\
		B_{y} &= h_5(a_1, a_2, a_3, d_2, d_3, \alpha_1, \alpha_2, R, z)\\
		C &= h_6(a_1, a_2, a_3, d_2, d_3, \alpha_1, \alpha_2, R, z) 
	\end{aligned}
\label{eq:conic_coeffs}
\end{equation}

\begin{figure}[H]
    \centering
    \begin{subfigure}{0.3\textwidth}
		\centering
		\includegraphics[width = \textwidth]{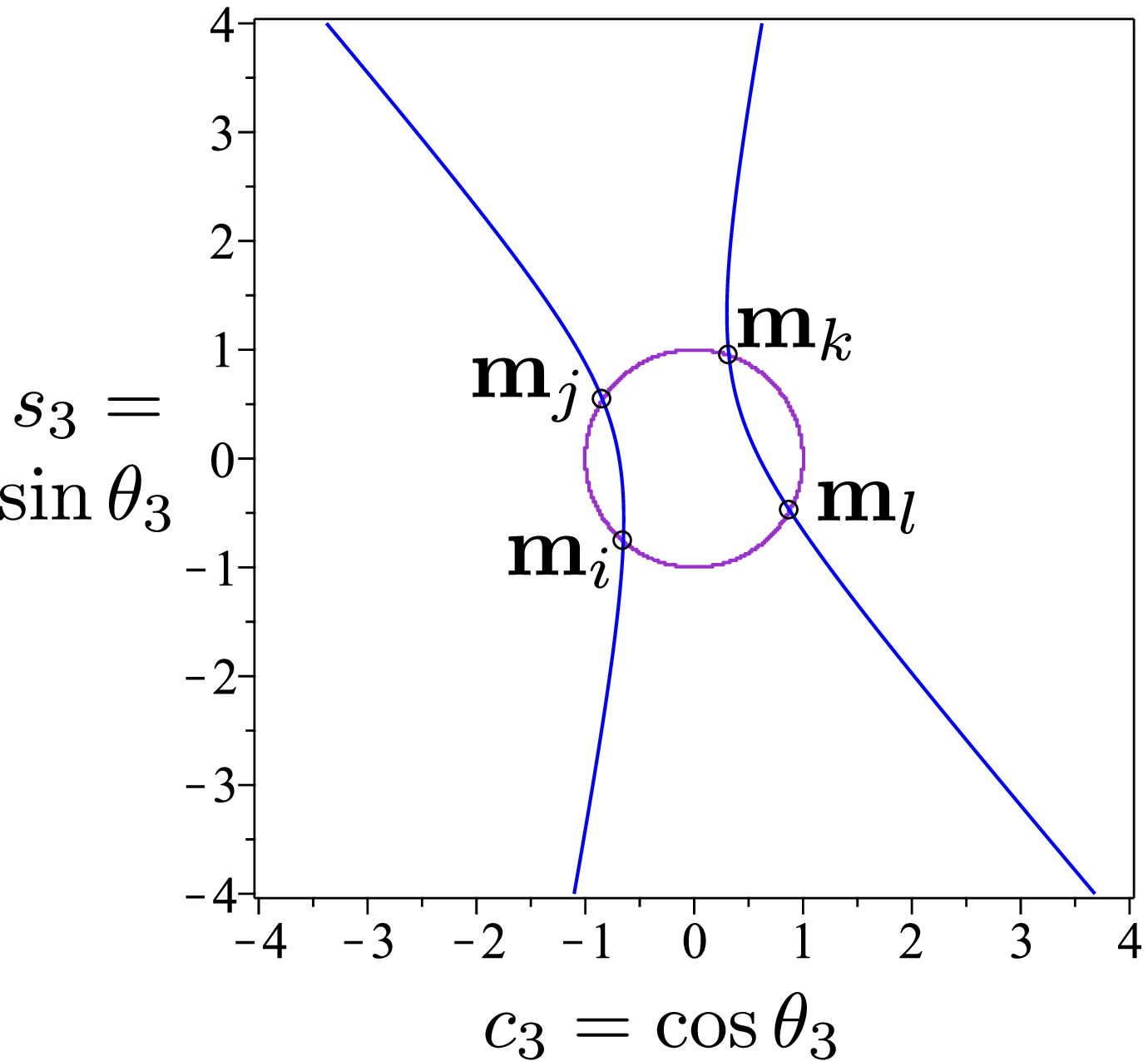}
		\caption{Hyperbola case}
		\label{fig:hyperbola}
	\end{subfigure}
	~
	\begin{subfigure}{0.3\textwidth}
		\centering
		\includegraphics[width = \textwidth]{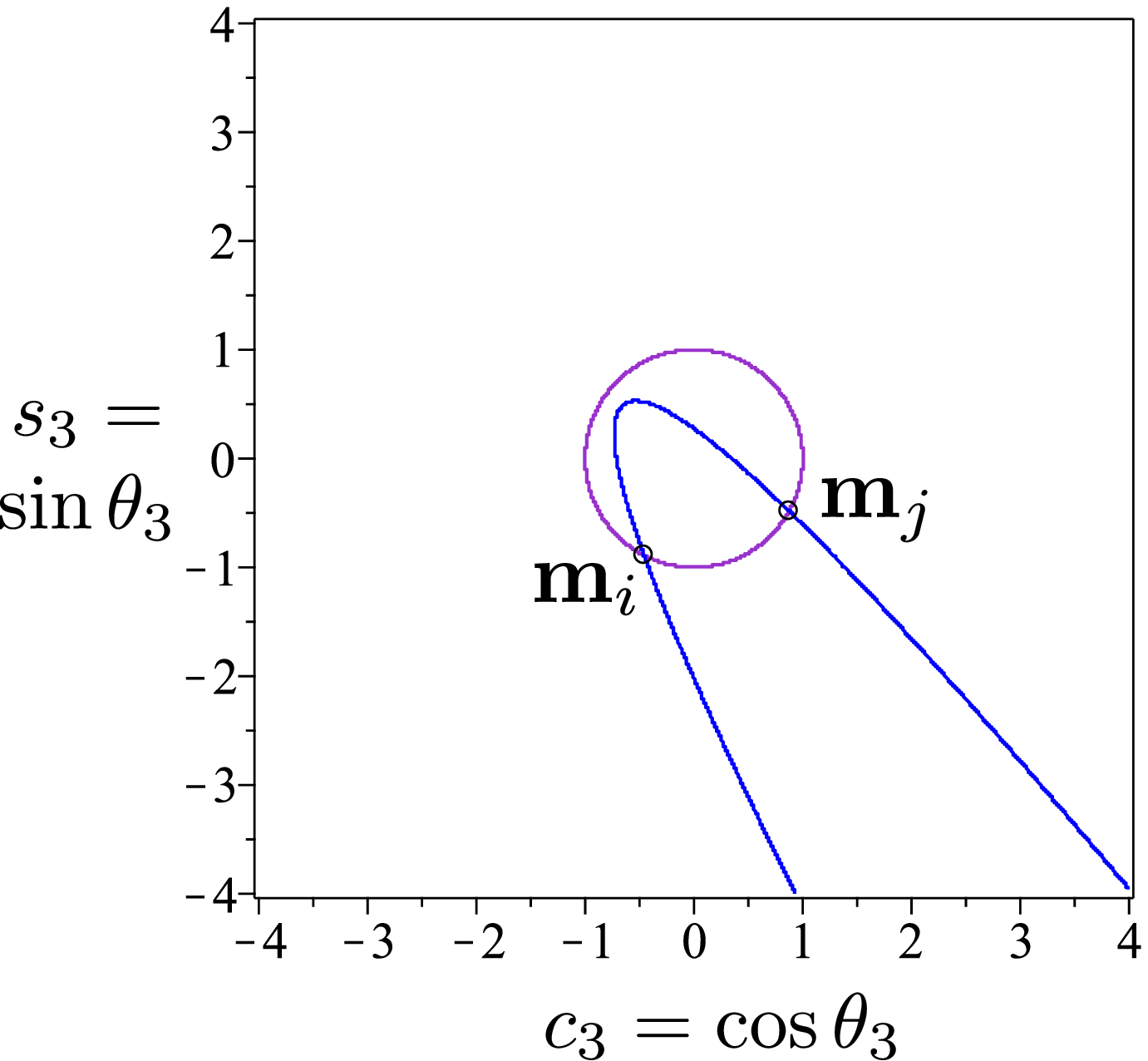}
		\caption{Parabola case}
		\label{fig:parabola}
	\end{subfigure}
	~
	\begin{subfigure}{0.3\textwidth}
		\centering
		\includegraphics[width = \textwidth]{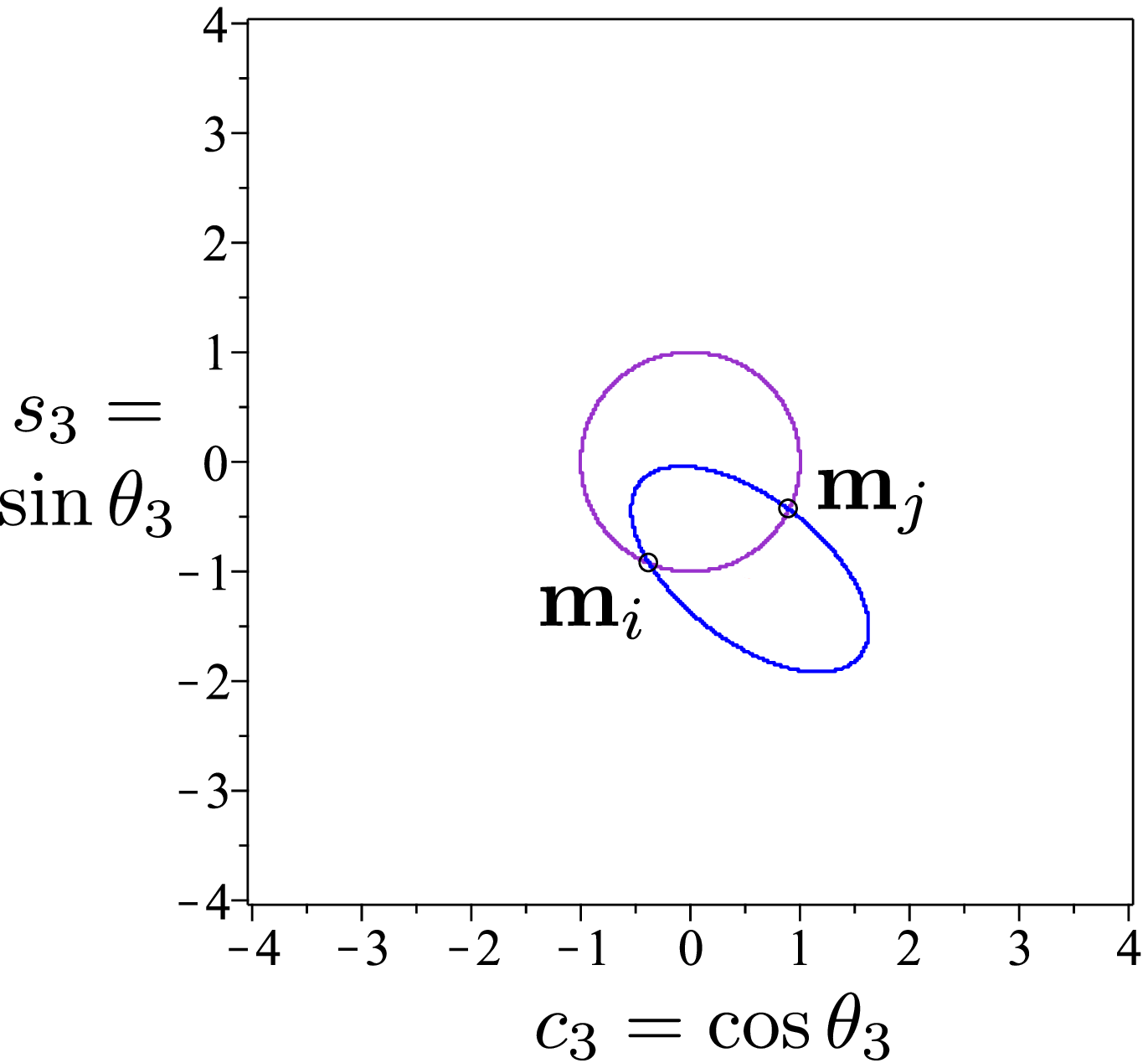}
		\caption{Ellipse case}
		\label{fig:ellipse}
	\end{subfigure}
    \caption{Intersection of the conic and unit circle in $c_3s_3$-plane for robots with different D-H parameters.\\ Robot parameters (\ref{fig:hyperbola}): d = [0, 1, 0], a = [1, 2, $\frac{3}{2}$], $\alpha$ = [$\frac{\pi}{2}$, $\frac{\pi}{6}$, 0], $(\rho, z) = (2.46, 0.15)$\\Robot parameters (\ref{fig:parabola}): d = [0, 1, 0], a = [1, 2, $\frac{3}{2}$], $\alpha$ = [$\frac{\pi}{3}$, $\frac{\pi}{2}$, 0], $(\rho, z) = (2.33, -0.26)$\\Robot parameters (\ref{fig:ellipse}): d = [0, 1, 0], a = [1, 2, $\frac{3}{2}$], $\alpha$ = [$\frac{\pi}{6}$,$\frac{\pi}{2}$,0], $(\rho, z) = (2.4, 0.6)$.}
    \label{fig:conic_all_case}
\end{figure}

The inverse kinematic solutions are defined by the intersection points between the conic (\ref{eq:conic_gen}) and the unit circle $c_3^2 + s_3^2 = 1$ in $c_3s_3$-plane. This conic can be a hyperbola, parabola or an ellipse depending on the D-H parameters and end-effector pose. An example of each one is shown in Fig. \ref{fig:conic_all_case}.

%d = [0, 1, 0] for all
%a = [1, 2, $\frac{3}{2}$], [1, 2, $\frac{3}{2}$], [4, 2, 6]
%alpha = [90, 30, 0], [60, 90, 0], [-90, 60, 0]
Performing the \changes{tangent half-angle} substitution, $t = \tan\frac{\theta_3}{2}$, we get a quartic inverse kinematic polynomial  $M(t) = at^4 + bt^3 + ct^2 + dt + e$ similar to the one mentioned in \cite{kohli_workspace_1985}. The coefficients of $M(t)$ are functions of the D-H parameters and of $R$ and $z$. The solutions to the polynomial equation, $M(t)=0$, are the intersection points between the conic and the circle and are labeled as $\mathbf{m}_{\psi}$, where $\psi \in \{i, j, k, l\}$ in the $c_3s_3$-plane.

\subsection{Singularities}
The Jacobian of $f$ at a certain configuration, denoted by $\mathbf{J(q)}$, is the Jacobian matrix of the robot at configuration $\mathbf{q}$: 

\begin{equation}
    \mathbf{J(q)}\ = \frac{\partial f(\mathbf{q})}{\partial \mathbf{q}}
    \label{eq:jac_def}
\end{equation}

The singularities are the critical points of $f$ in $\mathcal{J}$ and correspond to the set of all configurations in the joint space where the Jacobian matrix loses rank, i.e. when the determinant of $\mathbf{J}$ is zero. The critical values are the images of the critical points in $\mathcal{W}$. It is known that the roots of the inverse kinematic polynomial have multiplicity 2 or more at a singularity \cite{kohli_workspace_1985}. The algebraic expression of the singularity condition for an arbitrary 3R manipulator is recalled in Appendix A. The singularity in the workspace, the locus of critical value, is the image of the locus of critical points in the workspace and can be obtained from the inverse kinematic polynomial. The critical values in the workspace are those points where the following relation is satisfied:
\begin{align*}
    M(t) &= 0 \\
    \dfrac{\partial M(t)}{ \partial t} &= 0
\end{align*}
Where, $t = \tan \frac{\theta_3}{2}$ and $M(t)$ is the quartic inverse kinematic polynomial related to a 3R serial robot. The resulting algebraic expression is very large and is not reported here, see \cite{kohli_workspace_1985} and \cite{pseudo_cite} for more details.

\begin{figure}
	\centering
	\begin{subfigure}{0.45\textwidth}
		\centering
		\includegraphics[width = \textwidth]{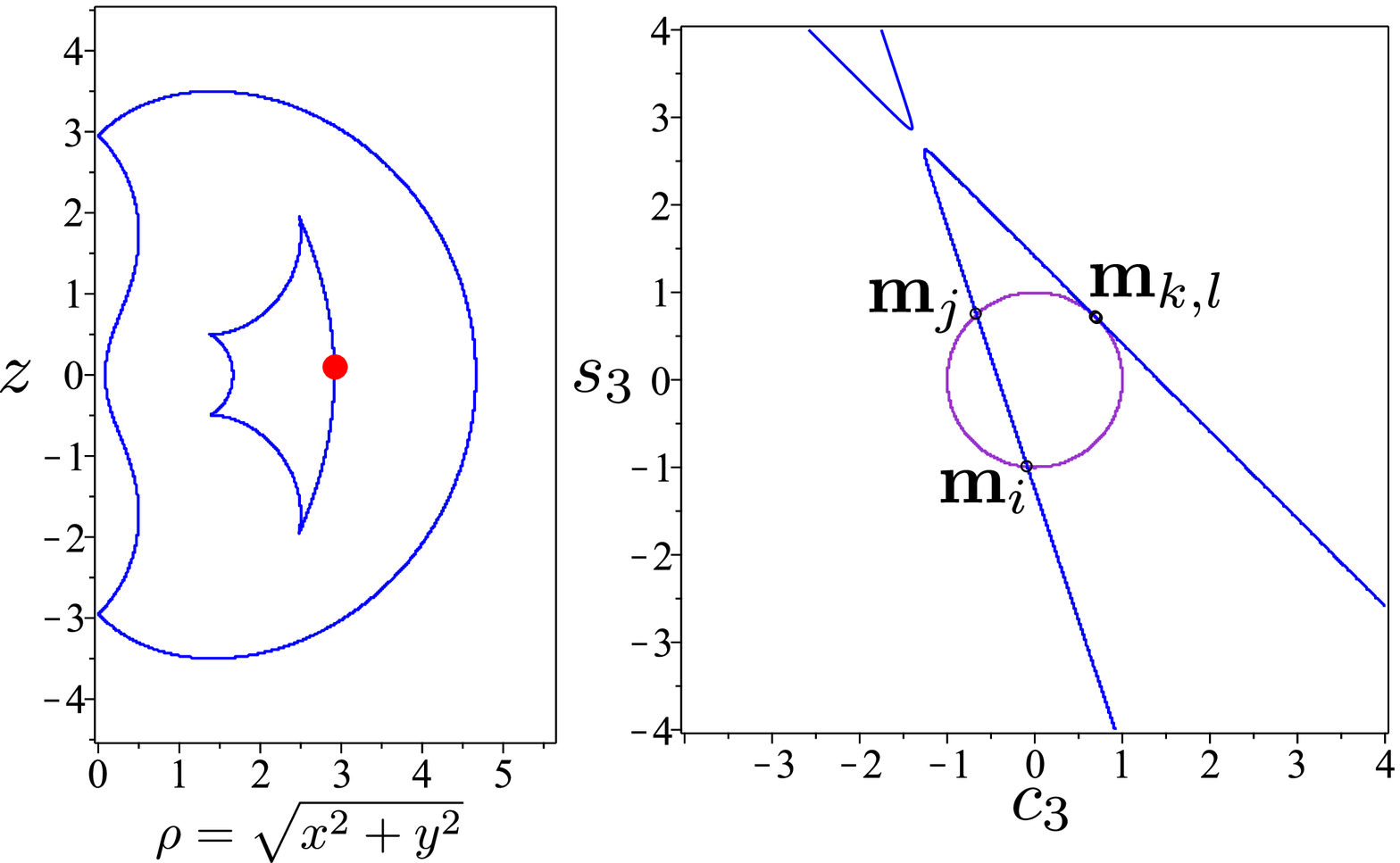}
		\caption{Point in workspace with root multiplicity 2}
		\label{fig:single_tangent}
	\end{subfigure}
	\\
	\begin{subfigure}{0.45\textwidth}
		\centering
		\includegraphics[width = \textwidth]{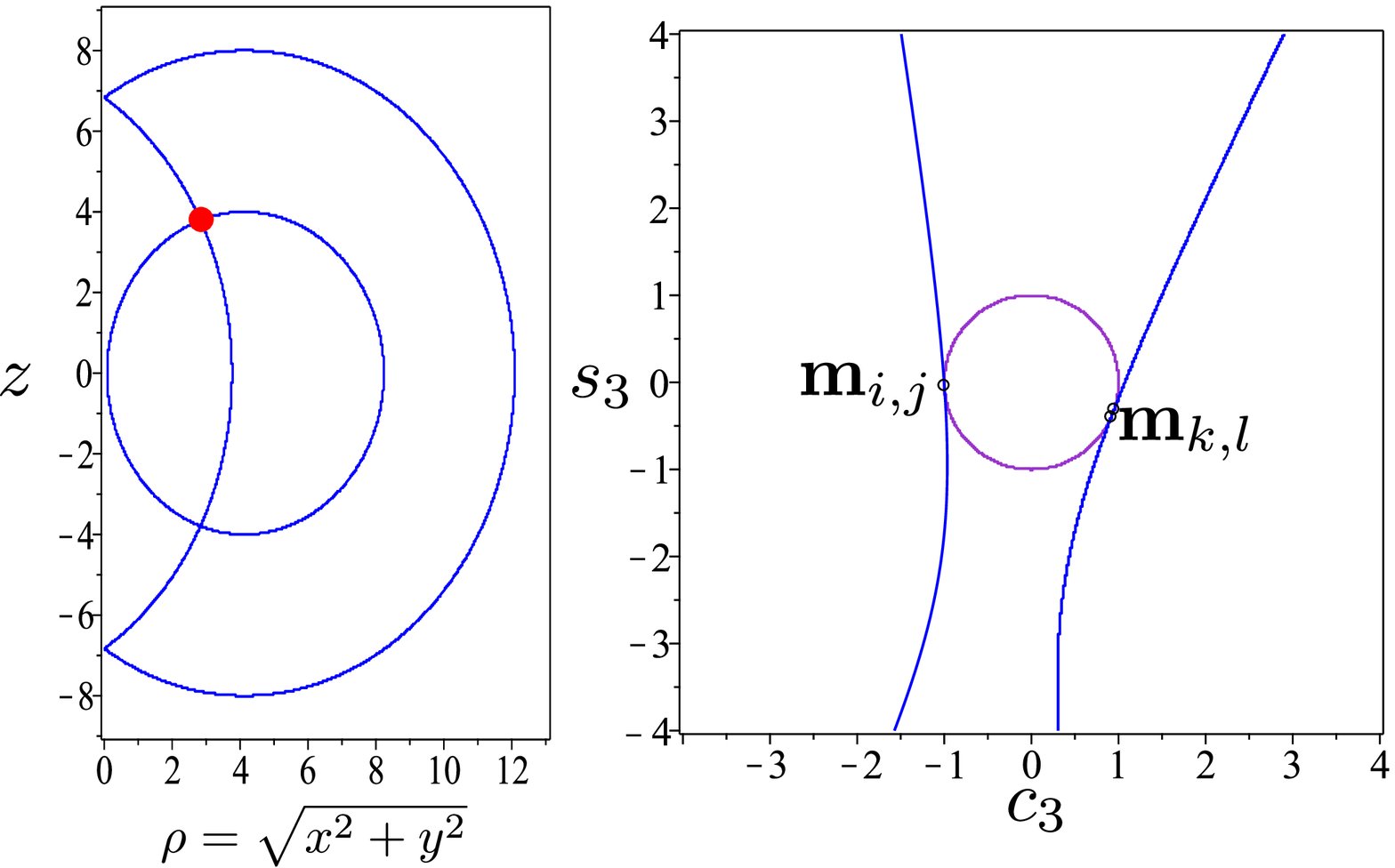}
		\caption{Node point - pair of roots with root multiplicity 2}
		\label{fig:double_tangent}
	\end{subfigure}
	\\
	\begin{subfigure}{0.45\textwidth}
		\centering
		\includegraphics[width = \textwidth]{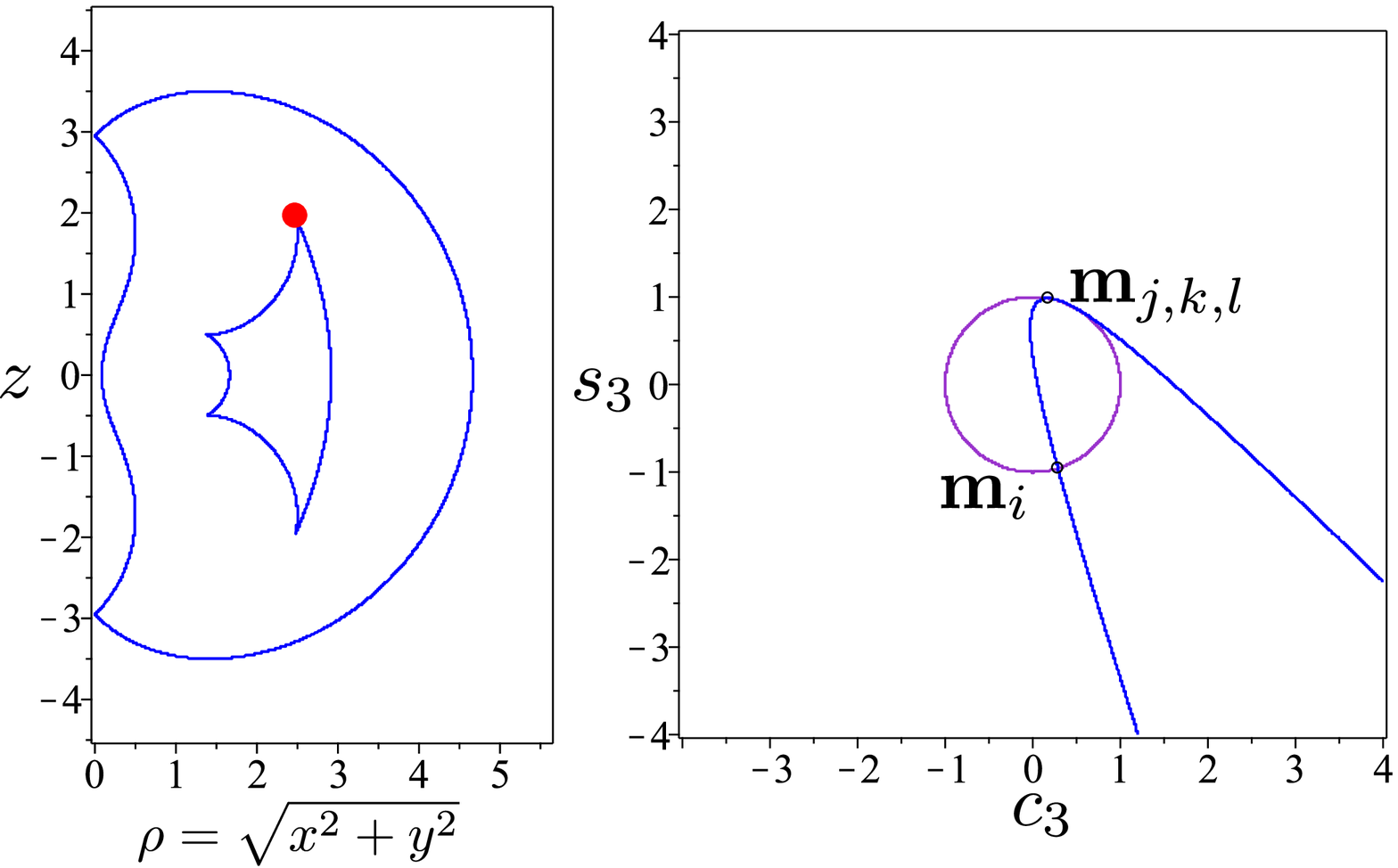}
		\caption{Cusp point in workspace with root multiplicity 3}
		\label{fig:cusp_tangent}
	\end{subfigure}
	\caption{Types of critical values in the workspace and corresponding tangency in $c_3s_3$-plane. \\Robot parameters (\ref{fig:single_tangent}): d = [0, 1, 0], a = [1, 2, $\frac{3}{2}$], $\alpha$ = [-$\frac{\pi}{2}$, $\frac{\pi}{2}$, 0], $(\rho, z) = (2.913, 0.1)$.\\
		Robot parameters (\ref{fig:double_tangent}): d = [0, 1, 0], a = [4, 2, 6], $\alpha$ = [-$\frac{\pi}{2}$, $\frac{\pi}{2}$, 0], $(\rho, z) = (2.84, 3.79)$ \\Robot parameters (\ref{fig:cusp_tangent}): d = [0, 1, 0], a = [1, 2, $\frac{3}{2}$], $\alpha$ = [-$\frac{\pi}{2}$, $\frac{\pi}{2}$, 0], $(\rho, z) = (2.48, 1.96)$.}
	\label{fig:tangent_case}
\end{figure}

With the conic representation, the geometric interpretation of a singularity associated with a double root is a point where the conic is tangent to the circle, as shown in Fig. \ref{fig:tangent_case}. The geometrical interpretation of a singularity associated with a root multiplicity higher than 2 is discussed in \changes{detail} in \cite{smith_analysis_1990, smith_higher_1993}.

It is known that the singularities of 3R serial robots are independent of the first joint angle, $\theta_1$ \cite{pai_genericity_1992}. This allows one to reduce the 3-dimensional joint space to ($\theta_2$, $\theta_3$). Consequently, the workspace is symmetric about the first joint axis. Assuming unlimited joints, it can thus be described by a \changes{half-cross section} in the plane ($\rho = \sqrt{x^2 + y^2}$, $z$). 

\subsection{Recall of important definitions}
%\begin{defin}
	\emph{Generic 3R serial robot}: A 3R serial robot is generic if and only if there exists only rank-2 singularities, i.e., the locus of critical points in joint space has no self-intersection or does not include any isolated point singularity \cite{pai_genericity_1992}.
	%\label{def:gen_rob}
%\end{defin}

\begin{figure}
    \centering
    \includegraphics[width=0.4\textwidth]{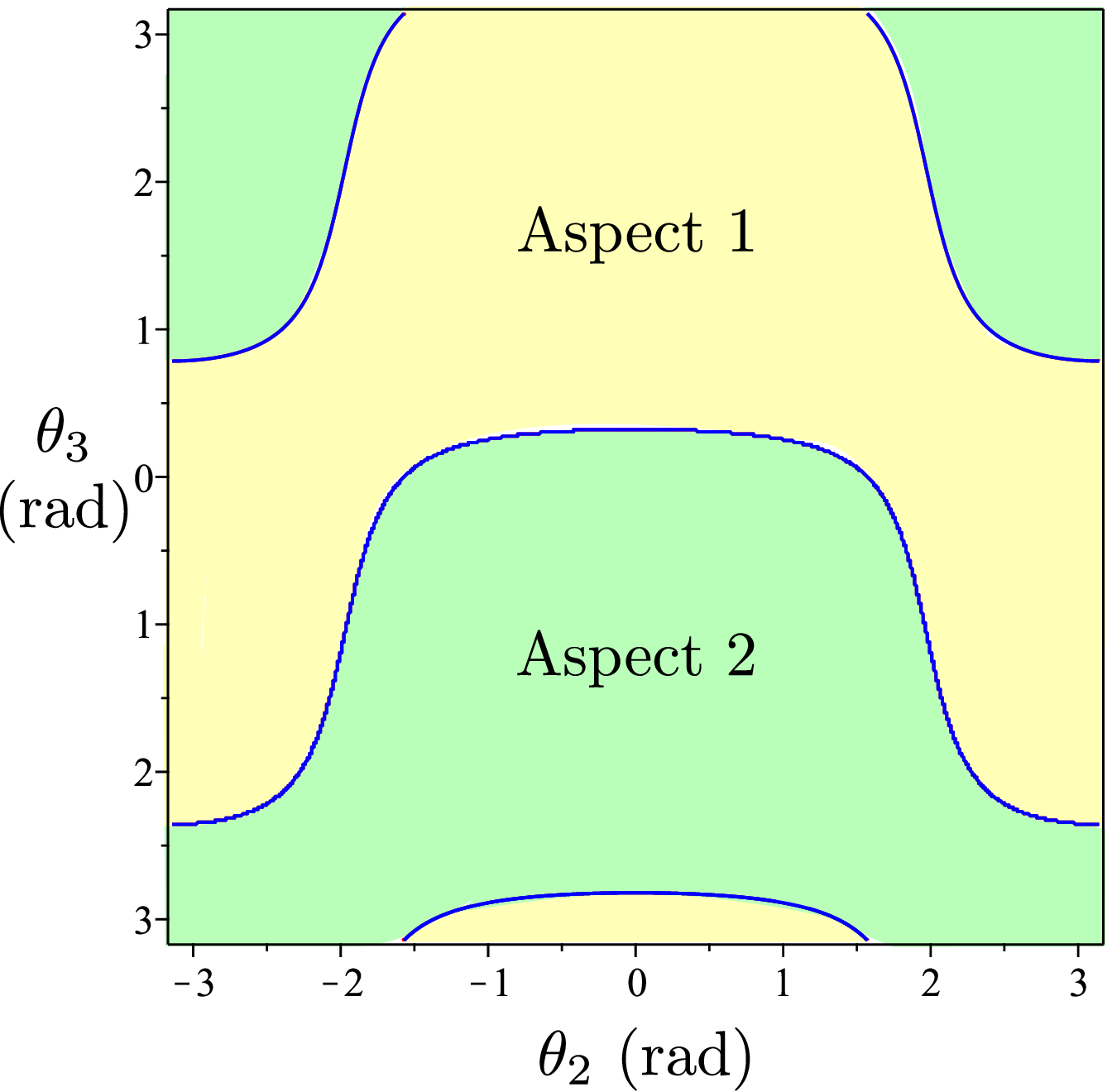}
    \caption{The two singularity free connected regions, called aspects, in joint space for a 3R serial robot \changes{\\Robot parameters: d = [0, 1, 0], a = [1, 2, $\frac{3}{2}$], $\alpha$ = [-$\frac{\pi}{2}$, $\frac{\pi}{2}$, 0]}}
    \label{fig:aspects}
\end{figure}

%\begin{defin}
	\emph{Aspects}: The aspects are the largest singularity free connected regions in the joint space of a serial robot \cite{borrel_study_1986}.  Figure \ref{fig:aspects} shows two aspects in the joint space of \changes{a 3R serial robot}.
	%\label{def:aspects}
%\end{defin}

%\begin{defin}
\emph{Cusp}: A cusp is a point in the workspace of a serial robot that satisfies the following conditions \cite{wenger_cuspidal_2019}:

\begin{equation}
	\begin{cases}
		&M(t) = 0\\
		&\frac{\partial M}{\partial t}(t) = 0 \\
		&\frac{\partial^2 M}{\partial t^2}(t) = 0 \\
	\end{cases}
	\label{eq:cusp_point}
\end{equation}	
 where $M(t)$ is the inverse kinematic polynomial of degree four \changes{of} a generic 3R serial robot. In Figs. \ref{fig:nonsingular_traj}, \ref{fig:single_tangent} and \ref{fig:cusp_tangent}, the robot has four cusps located at the corners of the inner region.
%\label{def:triple_point}
%\end{defin}

 The cusp also has to satisfy: 
\begin{equation}
	\frac{\partial^3 M}{\partial t^3}(t) \neq 0
	\label{eq:inequality}
\end{equation}
in order to exclude quadruple roots. However, it was shown in \cite{pai_genericity_1992} that quadruple roots cannot exist in generic 3R robots, and the condition in \eqref{eq:inequality} is thus always satisfied here. So, in the context of \changes{generic 3R serial robots, a cusp in the workspace can be identified solely by condition} (\ref{eq:cusp_point}). 

\emph{Node}: A node is a point in the workspace of a 3R serial robot where the inverse kinematic polynomial, $M(t)$, admits two distinct roots of multiplicity two as illustrated in Fig. \ref{fig:double_tangent}.

\emph{Cuspidal robot}: A robot for which there exists a path in the joint space connecting two inverse kinematic solutions without crossing the locus of critical points\changes{,} is defined as a cuspidal robot.

\emph{Pseudosingularity curve}: If $S$ is the set of critical points in the joint space, the pre-image of the critical values excluding $S$ is defined as the pseudosingularity curve \cite{pseudo_cite}:
	\begin{equation}
		PS = f^{-1}(f(S))\setminus S
		\label{eqn:PS}
	\end{equation}	
	
\begin{figure}[htbp]
	\centering
	\includegraphics[width = 0.85\textwidth]{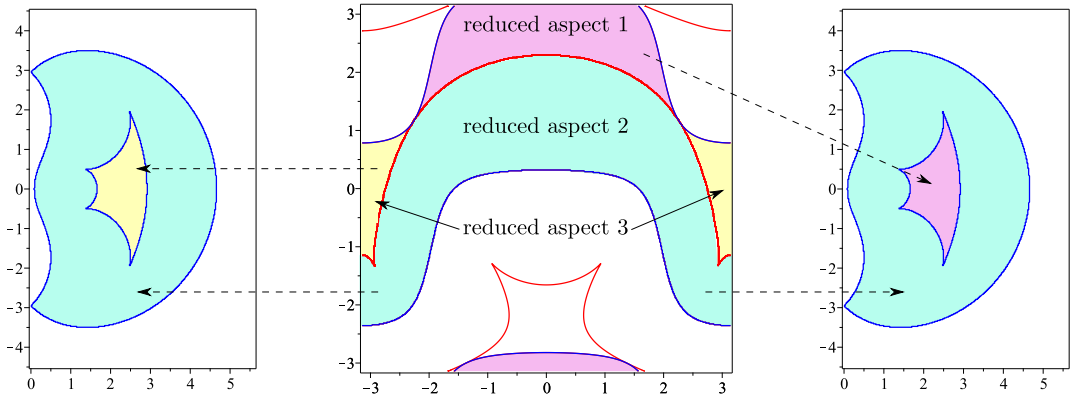}
	\caption{An example showing a set of reduced aspects present in an aspect of the joint space.\\	Robot parameters: d = [0, 1, 0], a = [1, 2, $\frac{3}{2}$], $\alpha$ = [$-\frac{\pi}{2}$, $\frac{\pi}{2}$, 0].}
	\label{fig:UDE}
\end{figure}

\emph{Reduced aspect}:	A reduced aspect is a region in the joint space that is bounded by the pseudosingularity curve and/or the locus of critical points and which has a one-to-one map to a bounded region in the workspace \cite{wenger_uniqueness_2004}. Fig. \ref{fig:UDE} illustrates an example of a set of reduced aspects in an aspect of the joint space for an orthogonal 3R cuspidal robot. The blue lines are the locus of critical points and critical values in the joint space and the workspace, respectively, while the red lines are the pseudosingularities present in the joint space. Note that the reduced aspects 1 and 3 in the joint space map to the same region in the workspace, suggesting two IKS in an aspect. 
    
\subsection{Nonsingular change of posture}
In the presented work, the joints of the robots are \changes{unlimited}, and thus the workspace is not constrained by the joint limits. A generic 3R robot may have up to 4 IKS at a given end-effector pose. An IKS can be defined by a point in the joint space, and a nonsingular change of posture can be described by a connected path between two IKS that does not cross the locus of critical points in the joint space. %\changes{Figure \ref{fig:nspc_wscs} shows a of nonsingular change of posture of an orthogonal robot.} 

\begin{figure}
	\centering
	\begin{subfigure}{0.8\textwidth}
		\centering
		\includegraphics[width = \textwidth]{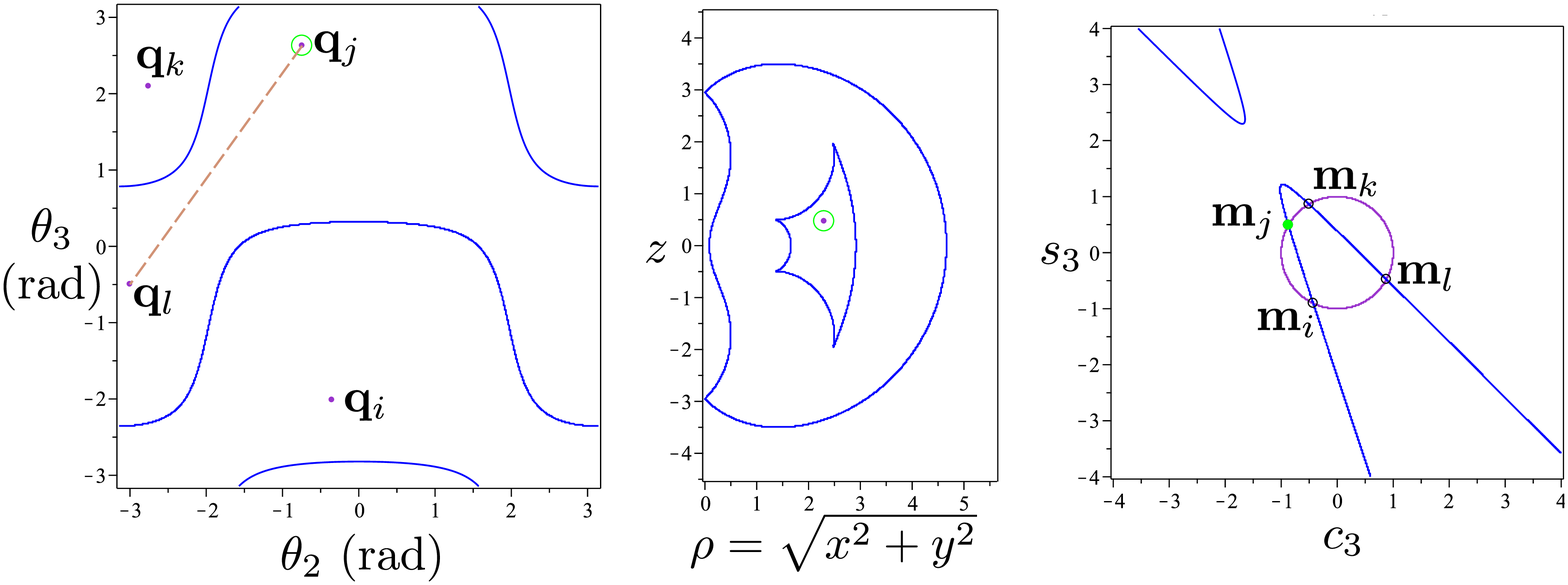}
		\caption{Phase 1: Starting from a point in workspace with 4 IKS}
		\label{fig:phase1}
	\end{subfigure}
	\\
	\begin{subfigure}{0.8\textwidth}
		\centering
		\includegraphics[width = \textwidth]{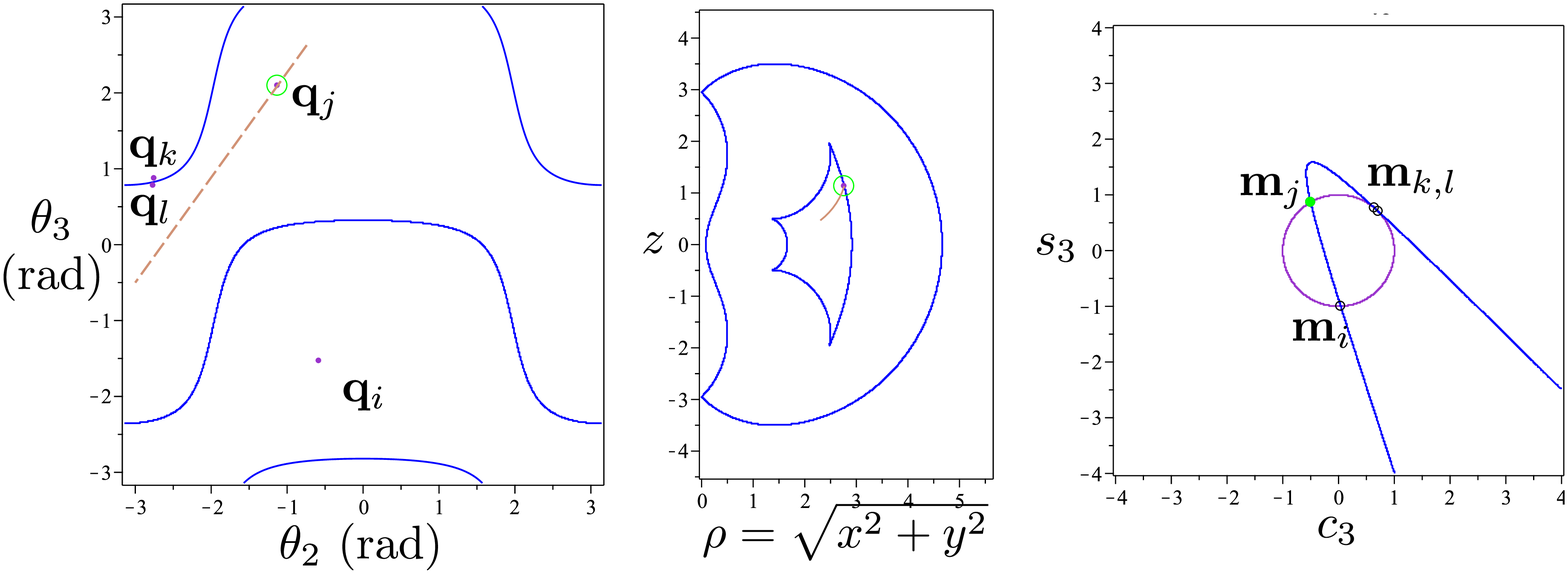}
		\caption{Phase 2: Entering a 2 solution region in workspace}
		\label{fig:phase2}
	\end{subfigure}
	\\
	\begin{subfigure}{0.8\textwidth}
		\centering
		\includegraphics[width = \textwidth]{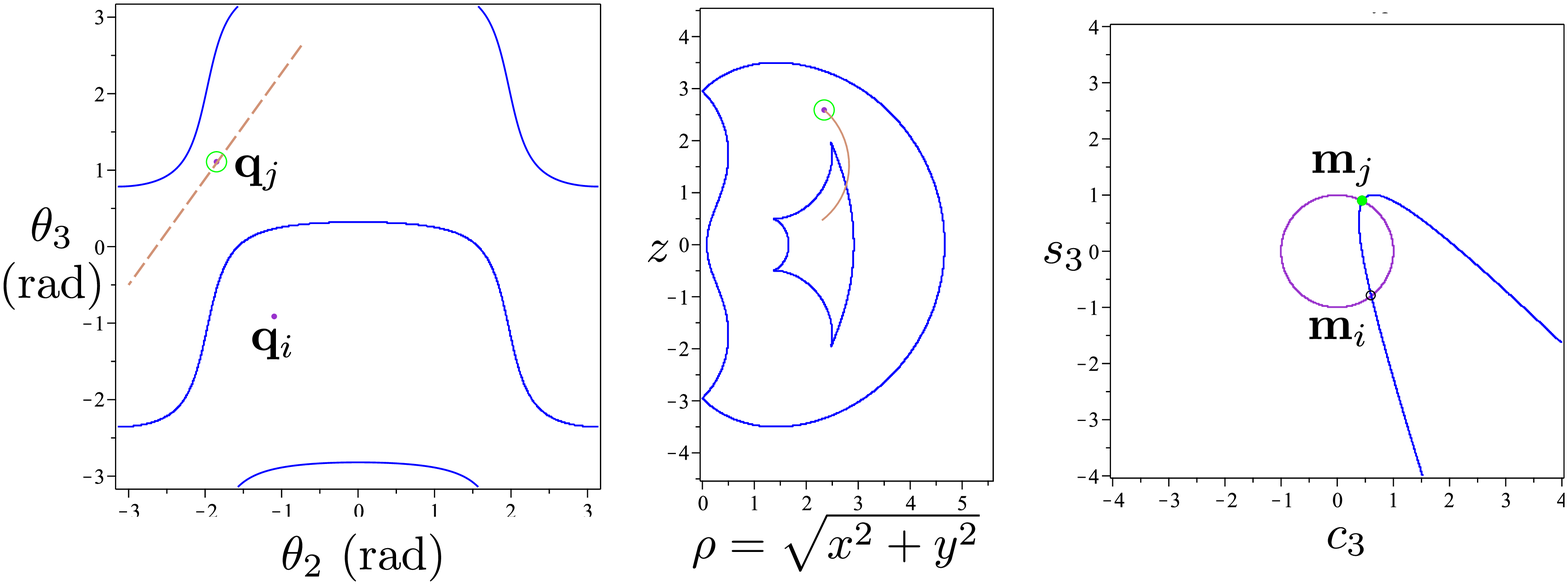}
		\caption{Phase 3: intersection point crosses the vertex of the conic }
		\label{fig:phase3}
	\end{subfigure}
	\\
	\begin{subfigure}{0.8\textwidth}
		\centering
		\includegraphics[width = \textwidth]{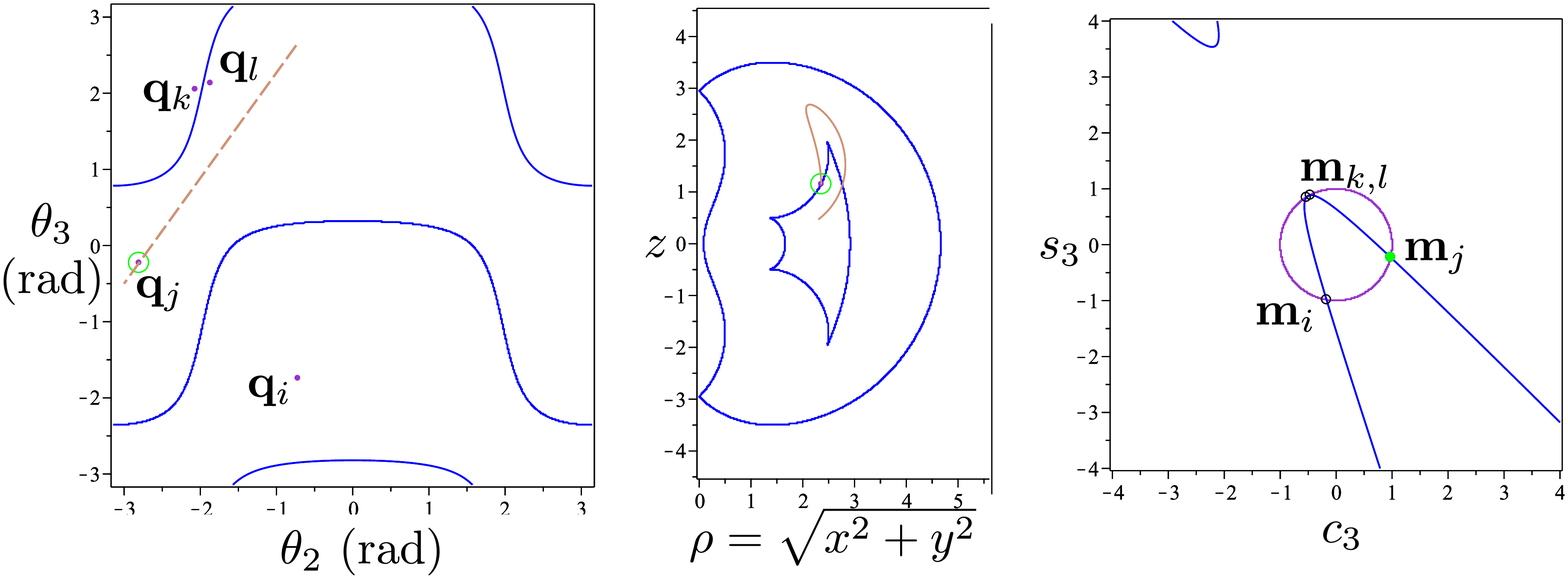}
		\caption{Phase 4: Re-entering the 4-solution region in workspace}
		\label{fig:phase4}
	\end{subfigure}
	\end{figure}%
    \begin{figure}[ht]\ContinuedFloat
    \centering
	\begin{subfigure}{0.8\textwidth}
		\centering
		\includegraphics[width = \textwidth]{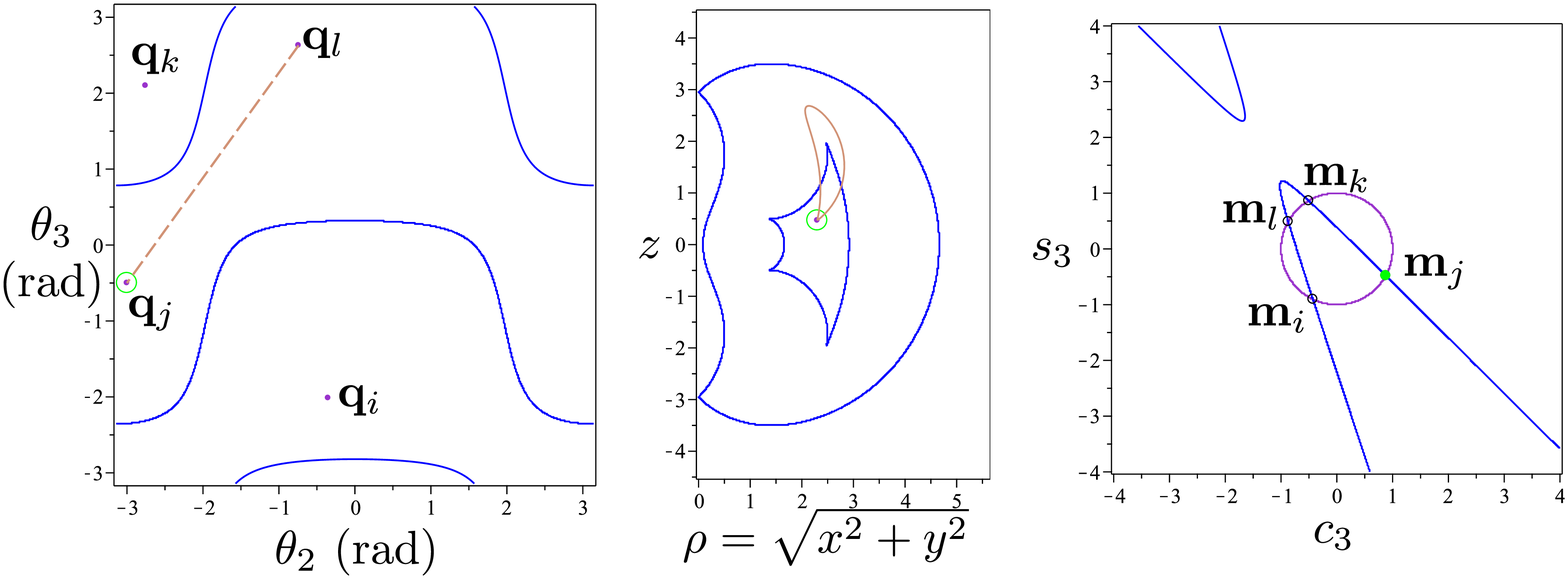}
		\caption{Phase 5: Reaching the same position in workspace}
		\label{fig:phase5}
	\end{subfigure}
	\captioncomment{An example of a nonsingular change of posture in the joint space and the workspace, and its corresponding geometrical interpretation in the $c_3s_3$-plane.}{Robot parameters: d = [0, 1, 0], a = [1, 2, $\frac{3}{2}$], $\alpha$ = [-$\frac{\pi}{2}$, $\frac{\pi}{2}$, 0].\\\changes{Path} in the joint space ($\theta_2$, $\theta_3$): from (-0.742, 2.628) to (-3, -0.5).}
	\label{fig:nspc_wscs}
\end{figure} 
In the workspace, a nonsingular change of posture defines a loop as we end up at the same position we started from. It has been noted in \cite{wenger_cuspidal_2019} that the nonsingular trajectory in workspace always starts from \changes{a point with four IKS}. The nature of this trajectory in the workspace will be studied in \changes{detail} in the coming sections. 

In the $c_3s_3$-plane, the nonsingular change of posture has an interesting interpretation. If we have four intersection points, $\mathbf{m}_i$, $\mathbf{m}_j$, $\mathbf{m}_k$ and $\mathbf{m}_l$,  between the conic and the unit circle in $c_3s_3$-plane corresponding to the four IKS at a particular end-effector pose, then the nonsingular change of posture between two IKS corresponding to $\mathbf{m}_j$ and $\mathbf{m}_l$ is such that $\mathbf{m}_j$ switches with $\mathbf{m}_l$ without vanishing as an intersection point of the conic and the unit circle. An example of a nonsingular change of posture is illustrated in Fig. \ref{fig:nspc_wscs}.

\begin{figure}[H]
    	\centering
    	\includegraphics[width = 0.3\textwidth]{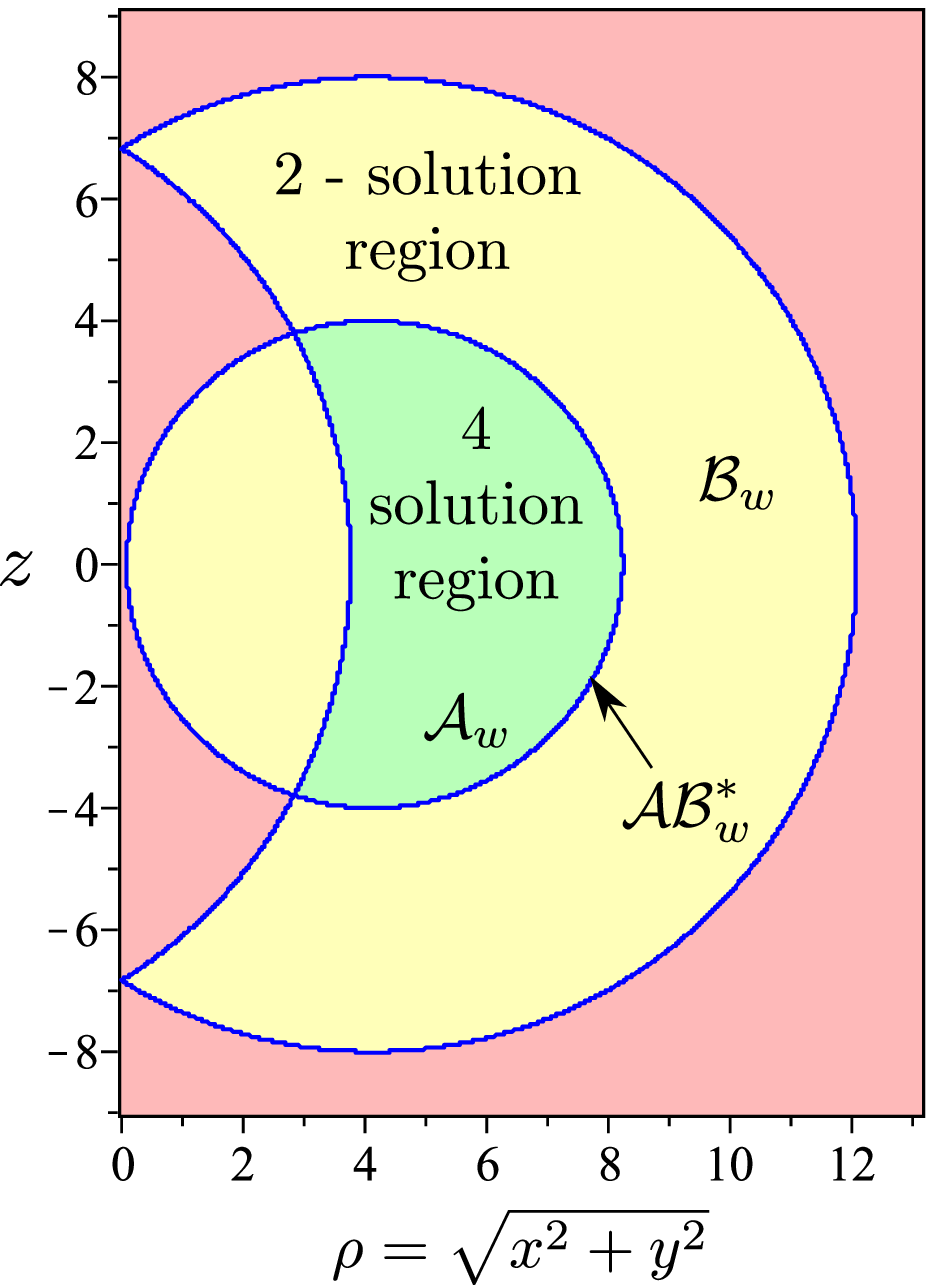}
    	\captioncomment{Regions separated by the locus \changes{of critical} values in the workspace. There are 3 IKS on $\mathcal{AB}_w^*$.}{Robot parameters: d = [0, 1, 0], a = [4, 2, 6], $\alpha$ = [-$\frac{\pi}{2}$, $\frac{\pi}{2}$, 0].}
    	\label{fig:wkspc_sep}	
    \end{figure}

\begin{figure}[H]
	\centering
	\includegraphics[width = 0.6\textwidth]{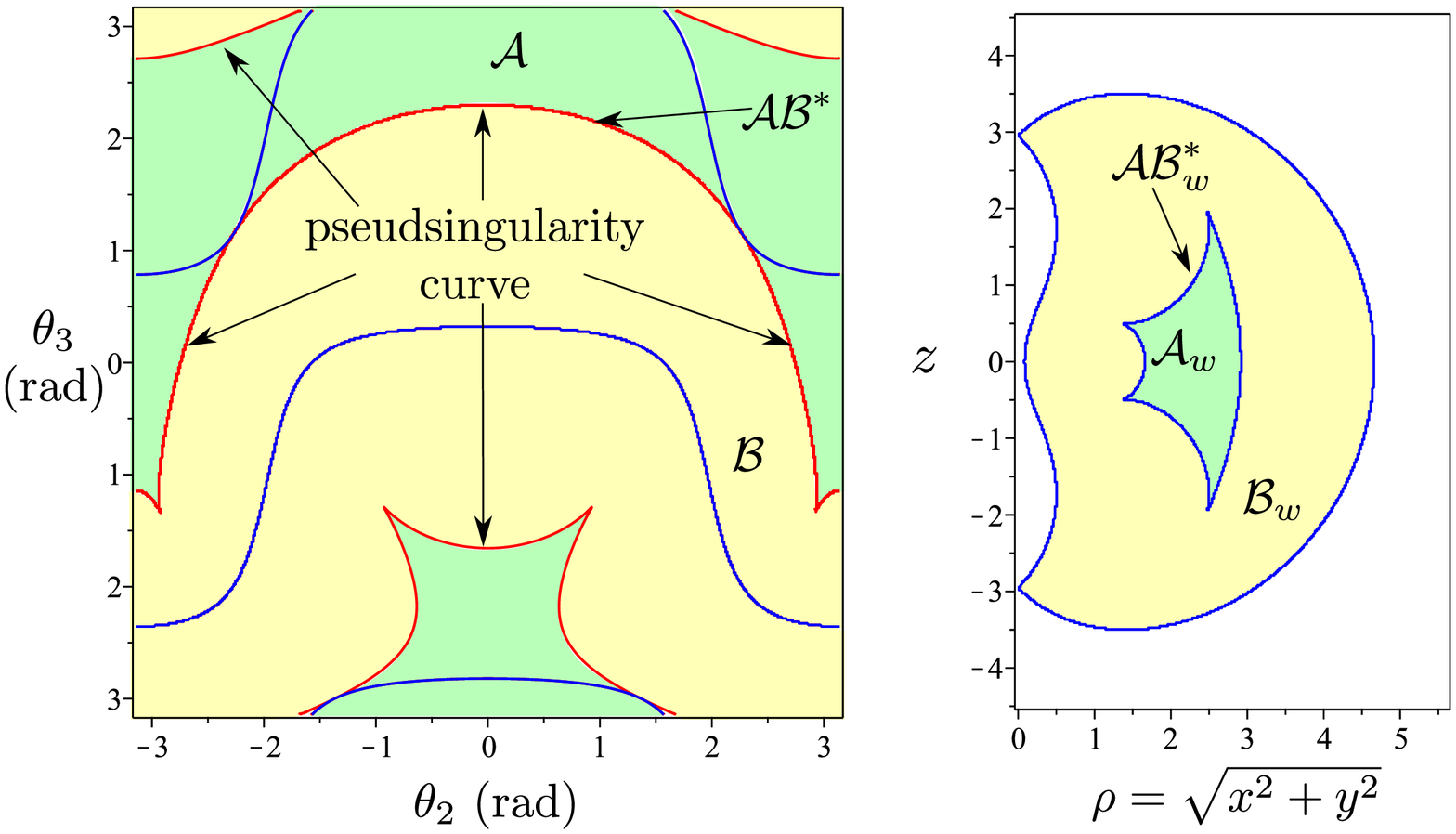}
	\captioncomment{An example of the regions separated by the pseudosingularity curve in joint space and the corresponding images in workspace.}{Robot parameters: d = [0, 1, 0], a = [1, 2, $\frac{3}{2}$], $\alpha$ = [-$\frac{\pi}{2}$, $\frac{\pi}{2}$, 0], ($\rho, z$) = (2.5, 0.5).}
	\label{fig:wspc_jspc}	
\end{figure}

\begin{propos}
	If $\mathcal{A}$ and $\mathcal{B}$ are two bounded regions in the same aspect sharing a common pseudosingularity curve $\mathcal{AB}^*$ and their image in the workspace belongs to regions $\mathcal{A}_w$ and $\mathcal{B}_w$ respectively, then the absolute difference between the number of IKS \changes{in region} $\mathcal{A}_w$ and \changes{in} region $\mathcal{B}_w$ is always two (refer to Fig. \ref{fig:wspc_jspc}). Moreover, the absolute difference between the number of IKS in region $\mathcal{A}_w$ or $\mathcal{B}_w$ and at any point on the boundary $\mathcal{AB}_w^*$ between them, is always one (Fig. \ref{fig:wkspc_sep}).
	\label{propos:jwsing}
\end{propos}

    This is a well-known property \cite{kohli_workspace_1985} and is commonly interpreted as two inverse kinematic solutions merge at a singular configuration. It is also important to note that for a generic 3R robot, the shared boundary does not include isolated finite points. 
    
    If a pseudosingularity exists in the joint space of a 3R serial robot, then each point on the pseudosingularity curve has an image on the locus of critical values in the workspace. Therefore, crossing a pseudosingularity curve in the joint space is similar to crossing the locus of critical values in the workspace, and thus the images of the regions sharing the pseudosingularity curve should have absolute difference of two. 

\subsection{Sufficient condition}    
\begin{figure}[H]
		\centering
		\includegraphics[width = 0.75\textwidth]{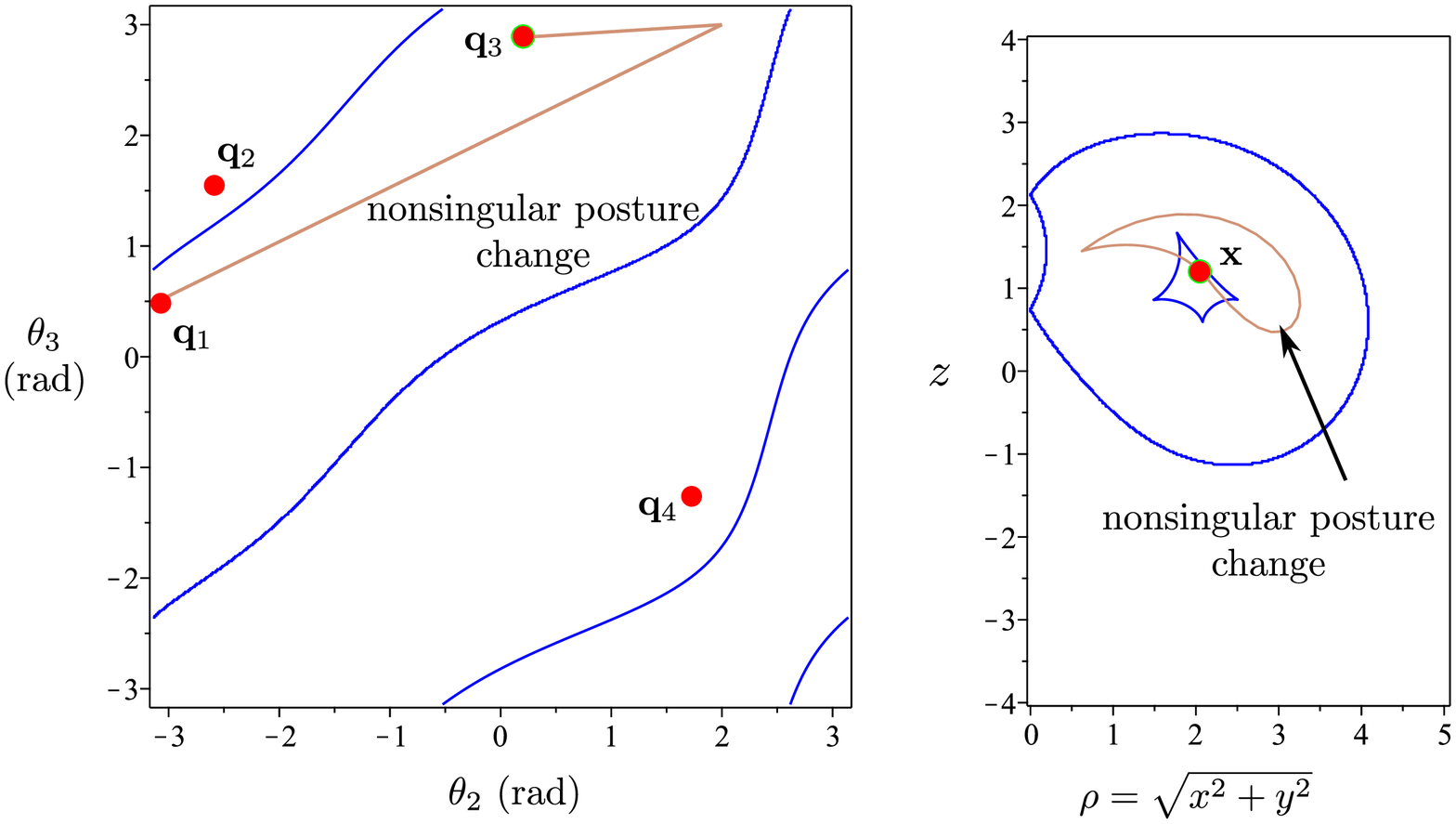}
		\caption{Example showing a non-local nonsingular change of posture in  joint space and workspace.\\Robot parameters: d = [0, 1, 0], a = [1, 2, 1], $\alpha = \left[\dfrac{\pi}{6}, \dfrac{\pi}{2}, 0\right]$.\\Trajectory in joint space ($\theta_2, \theta_3$): (-3, 0.5) to (2, 3) to (0.2, 2.8).}
		\label{fig:2cuspJSWS}
	\end{figure}

\begin{thm}
	\changes{The existence of a cusp in the workspace of a 3R robot} is a sufficient condition for the robot to be cuspidal.
	\label{propos:first}
\end{thm}
\begin{proof}
	Using Whitney's theorem \cite{whitney_singularities_1955}, it has been noted in \cite{corvez_study_2005}, that the existence of a cusp in the workspace of a 3R robot is equivalent to a nonsingular change of posture in a sufficiently small neighborhood of the cusp.
	\end{proof}
	It is important to note that the work in \cite{corvez_study_2005} does not establish the necessary and sufficient cuspidality condition, as the existence of a cusp can be confirmed only if we have a nonsingular change of posture in a \textit{sufficiently small} neighborhood. In Fig. \ref{fig:2cuspJSWS}, the nonsingular change of posture from $\mathbf{q}_1$ to $\mathbf{q}_3$ for a point, $\mathbf{x}$, in the workspace is not local and thus the equivalence in \cite{corvez_study_2005} cannot be used to prove that the above sufficient condition is also necessary. In the next section, we establish a proof that the existence of a cusp is also a necessary condition for a any generic 3R robot to be cuspidal.
 
 %%%%%%%%%%% section: conclusions %%%%%%%%%%%
 \section{Proof}
\label{section:proof}  
\begin{figure}
    \centering
    \includegraphics[width=0.7\textwidth]{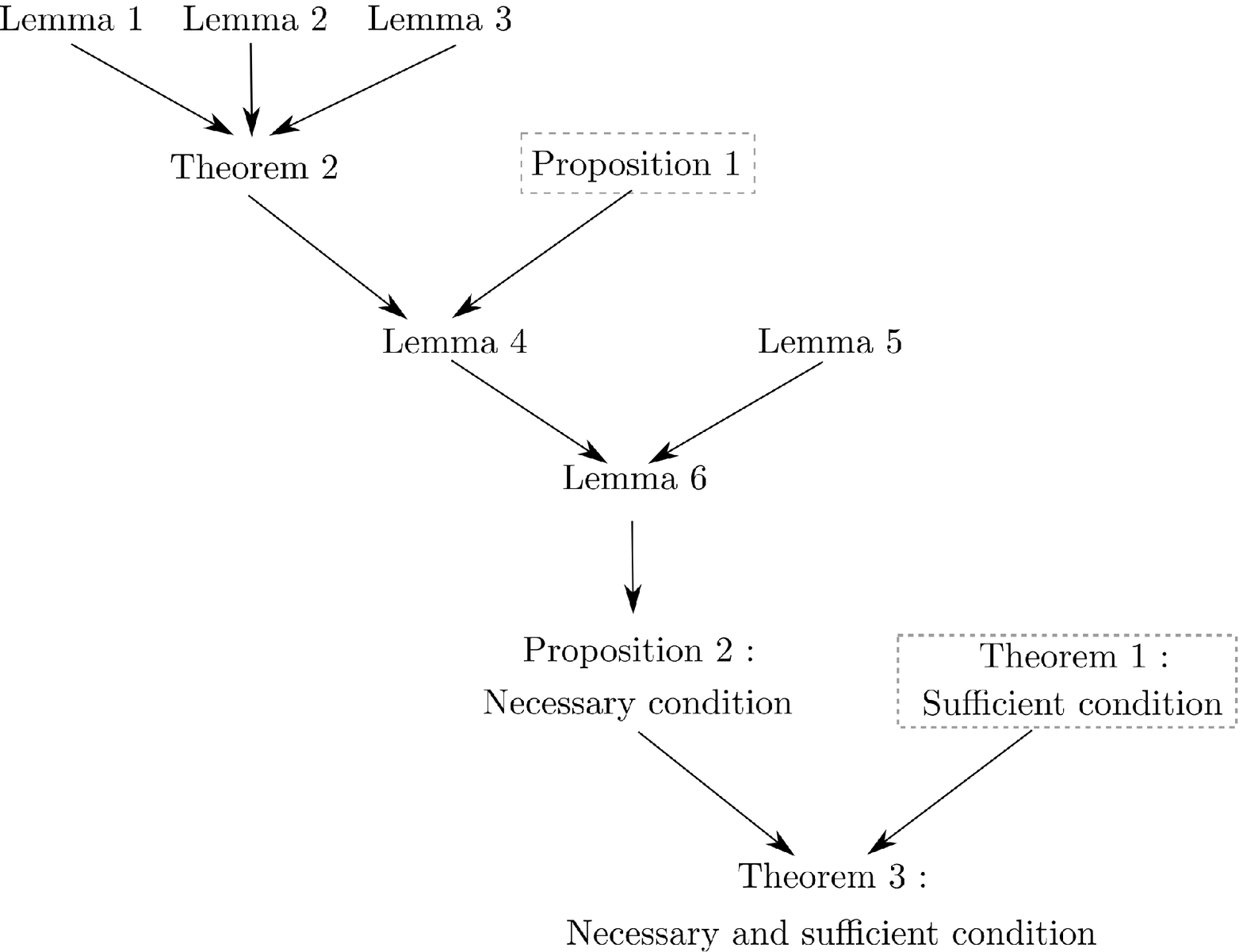}
    \caption{The flowchart of the proof for the necessary and sufficient condition.  Proposition 1 and Theorem 1 are already proven and mentioned in Section \ref{section:preliminaries}.}
    \label{fig:flowchart}
\end{figure}

In this section, we present the proof of the necessary cuspidality condition for any generic \changes{3R robot}. The proof discusses several lemmas and uses proposition 1 and definitions in Section \ref{section:preliminaries} to arrive at a conclusion. Figure \ref{fig:flowchart} shows a flowchart illustrating the organization of the proof. This fowchart should be read as follows: Proposition 2 is the cuspidality necessary condition which, along with the cuspidality sufficient condition (Theorem 1), makes it possible to establish the necessary and sufficient condition of cuspidality in the end of this section (Theorem 3). To prove Proposition 2, Lemmas 1, 2 and 3 are first established, leading to Theorem 2 which, along with Proposition 1, leads to Lemma 4. Lemmas 4 and 5 lead to Lemma 6, which makes it possible to prove Proposition 2. 

\subsection{Necessary condition}
\begin{propos}
	If a generic 3R robot is cuspidal, then there exists a cusp in the workspace of \changes{this} robot.
	\label{propos:second}
\end{propos}

%We want to provide a proof by contradiction of the above proposition. 
We prove Proposition 2 by contradiction: we consider a hypothetical cuspidal robot that has no cusps in its workspace and \changes{we} show that this case cannot exist. In order to do so, we first need to set and prove a series of lemmas.
\begin{lemma}
	The nature of the conic, i.e. ellipse, hyperbola or parabola, related to a particular set of \changes{the} D-H parameters of a generic 3R robot remains the same throughout the workspace of the robot.
	\label{lemma:conic_nature}
\end{lemma}
\begin{proof}
	The determinant of the matrix $\mathbf{D}$ displayed in (\ref{eq:D_conic}) determines the nature of the conic. Since $A_{xx}, A_{xy}$ and $A_{yy}$ are functions of D-H parameters only as shown in (\ref{eq:conic_coeffs}), the nature of the conic remains the same throughout the workspace. 
	\begin{equation}
		\mathbf{D} = \begin{bmatrix} A_{xx} & A_{xy} \\ A_{xy} & A_{yy}\end{bmatrix}
		\label{eq:D_conic}
	\end{equation}
\end{proof}

\begin{lemma}
	The orientation of the principal axes of the conic related to a particular set of D-H parameters of a generic 3R serial robot is constant throughout the workspace of the robot.
	\label{lemma:conic_orientation}
\end{lemma}
\begin{proof}
	The eigenvectors of $\mathbf{D}$ determine the orientation of the principal axes and, as noted in the proof of Lemma \ref{lemma:conic_nature}, $\mathbf{D}$ is independent of $R$ and $z$ and the eigenvectors are thus constant for a given robot. 
\end{proof}

For a given point $\mathbf{p}$ in the workspace of a generic 3R robot, \changes{let $n$ $\in \{1, 2, 3, 4\}$ be} distinct preimages such that we have $n$ intersection points between the conic and the circle in $c_3s_3$-plane. We will say that an intersection point, $\mathbf{m}_j$, in $c_3s_3$-plane is adjacent to another intersection point, $\mathbf{m}_i$, if it is the first intersection point encountered after travelling in either clockwise or counterclockwise direction starting from $\mathbf{m}_i$. It has been illustrated in Fig. \ref{fig:tangent_case} that any path in the workspace starting from $\mathbf{p}$ to any point on the boundary, results in at least 2 intersection points coming together at the tangent point in $c_3s_3$-plane. Accordingly, the following Lemma is set:
\begin{lemma}
	In a generic 3R robot, two intersection points, $\mathbf{m}_a$ and $\mathbf{m}_b$, meet at a tangent point corresponding to roots of the inverse kinematic polynomial with multiplicity two, only if they are adjacent to each other in the cyclic ordering of the intersection points in the $c_3s_3$ - plane.
\end{lemma}
\begin{proof}
	For two points $\mathbf{m}_a, \mathbf{m}_b$ to meet together, $\mathbf{m}_a$ should start travelling towards $\mathbf{m}_b$, or $\mathbf{m}_b$ should travel towards $\mathbf{m}_a$. If there exists an intersection point $\mathbf{m}_c$ between them, then $\mathbf{m}_a$ and $\mathbf{m}_b$ can meet at a tangent point only after either $\mathbf{m}_a$ or $\mathbf{m}_b$ meets $\mathbf{m}_c$ at a tangent point. A graphical illustration of the tangency between the adjacent points is given in Fig. \ref{anim:sing_traj}.
	%%%%%%%%%% New figure %%%%%%%%%%
	\begin{figure}[htbp]
		\centering
		\begin{subfigure}{0.67\textwidth}
			\centering
			\includegraphics[width=\textwidth]{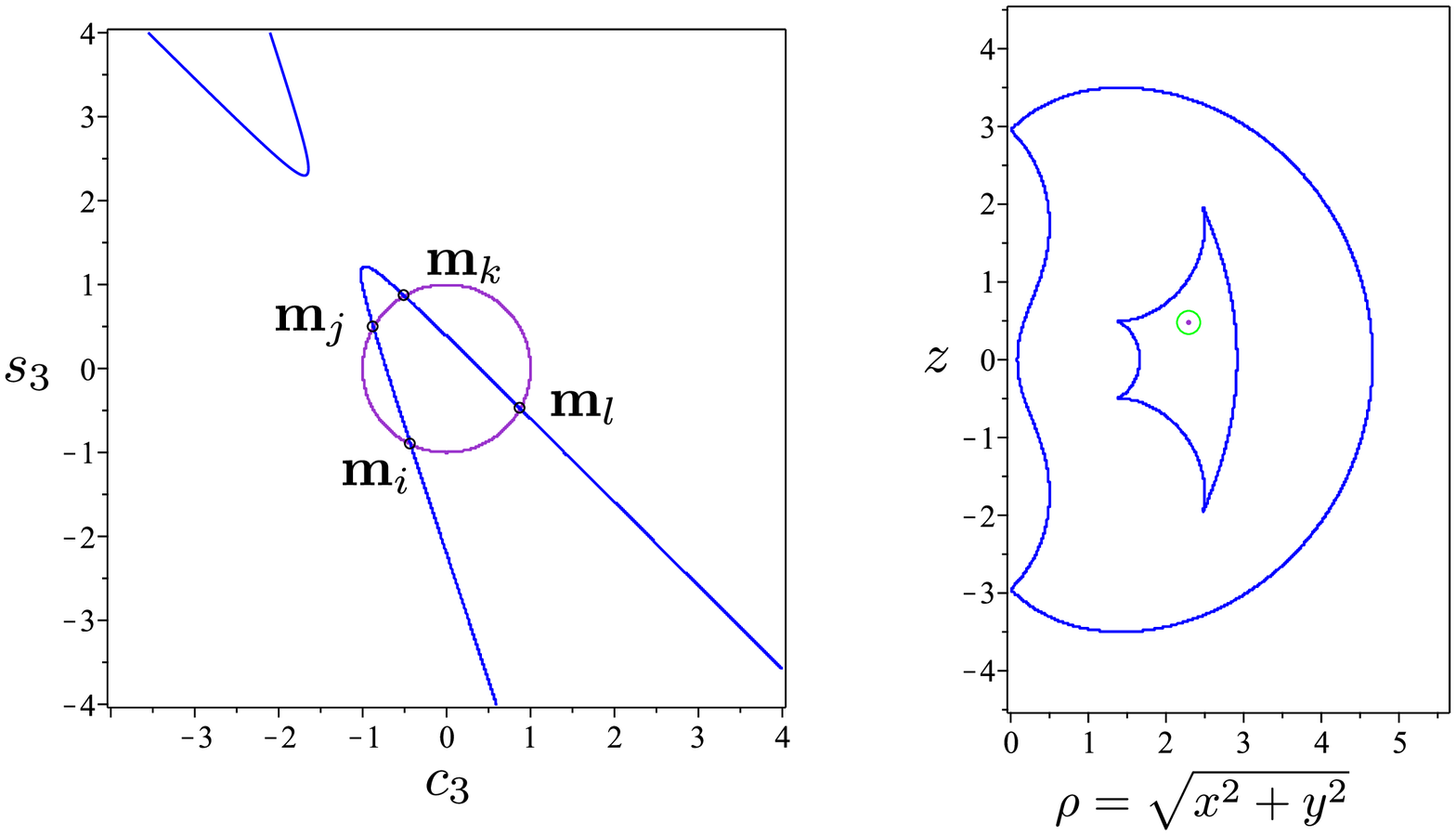}
			\caption{Initial configuration with four intersection points in $c_3s_3$ - plane}
			\label{fig:12a}
		\end{subfigure}
		\\
		\begin{subfigure}{0.67\textwidth}
			\centering
			\includegraphics[width=\textwidth]{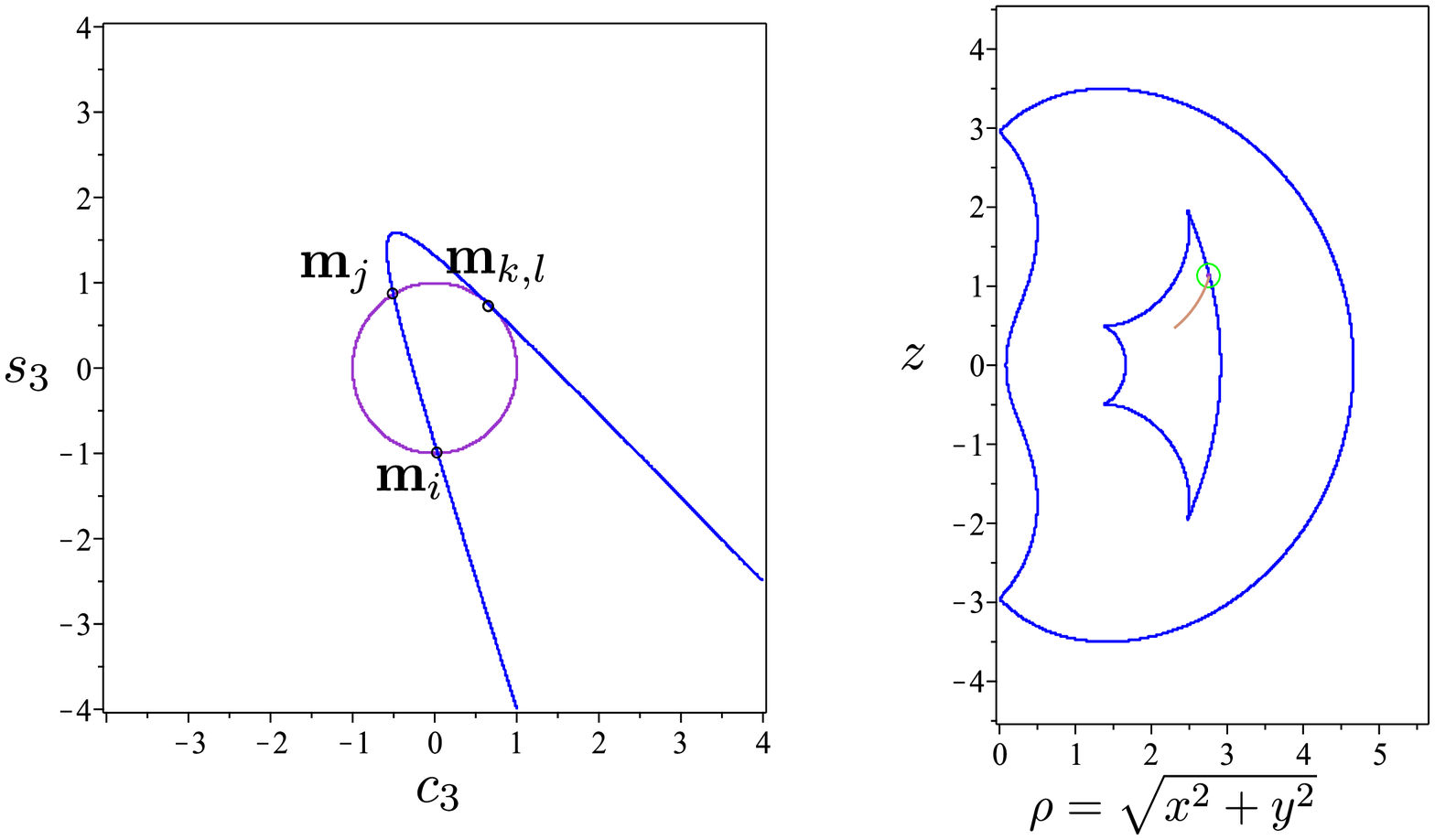}
			\caption{Tangency between adjacent points on either side of the vertex}
			\label{fig:12b}
		\end{subfigure}
		\\
		
		\begin{subfigure}{0.67\textwidth}
			\centering
			\includegraphics[width=\textwidth]{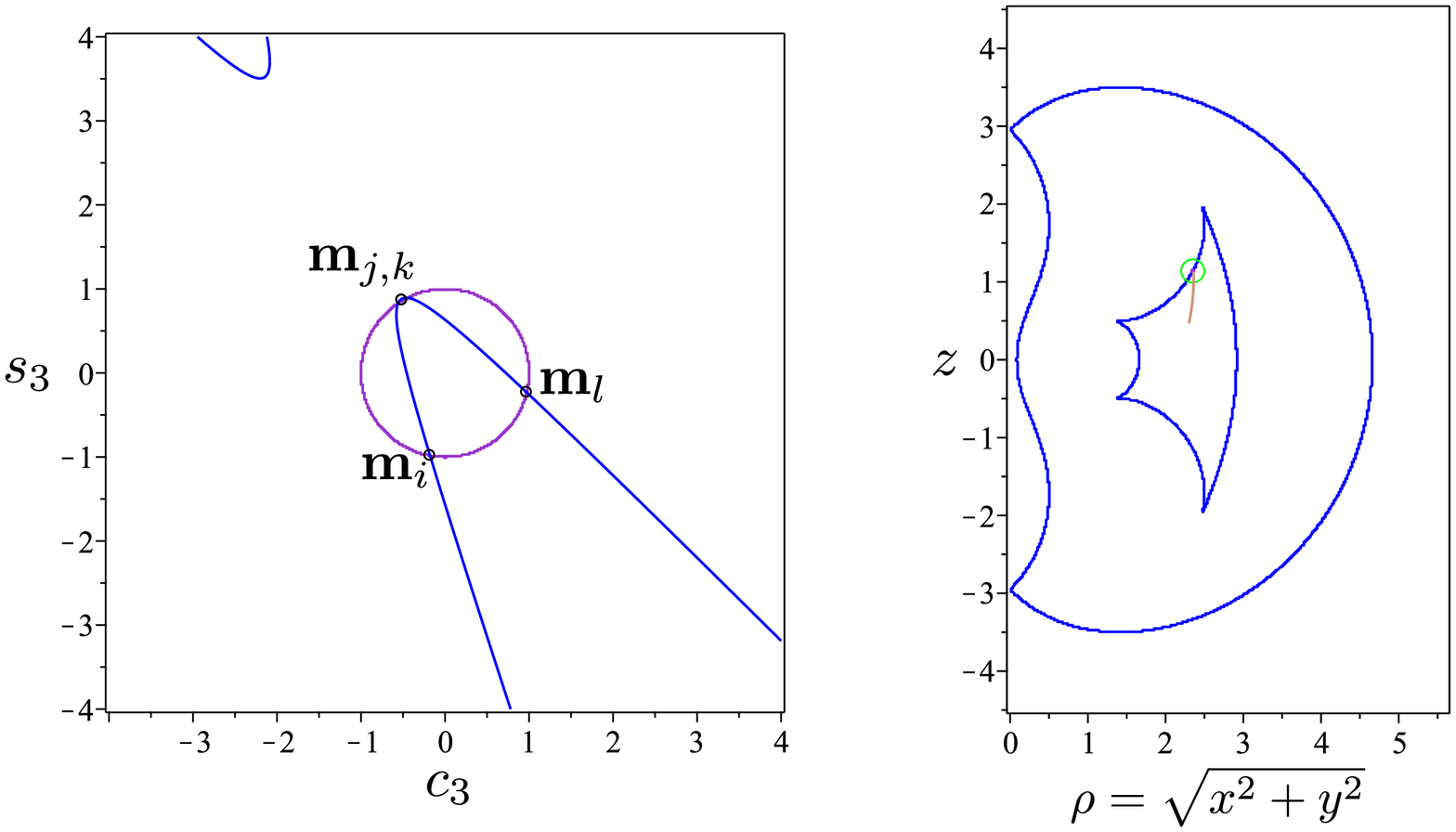}
			\caption{Tangency between adjacent points on same side of the vertex}
			\label{fig:12c}
		\end{subfigure}
		\end{figure}
		\begin{figure}\ContinuedFloat
		\centering
		\begin{subfigure}{0.7\textwidth}
			\centering
			\includegraphics[width=\textwidth]{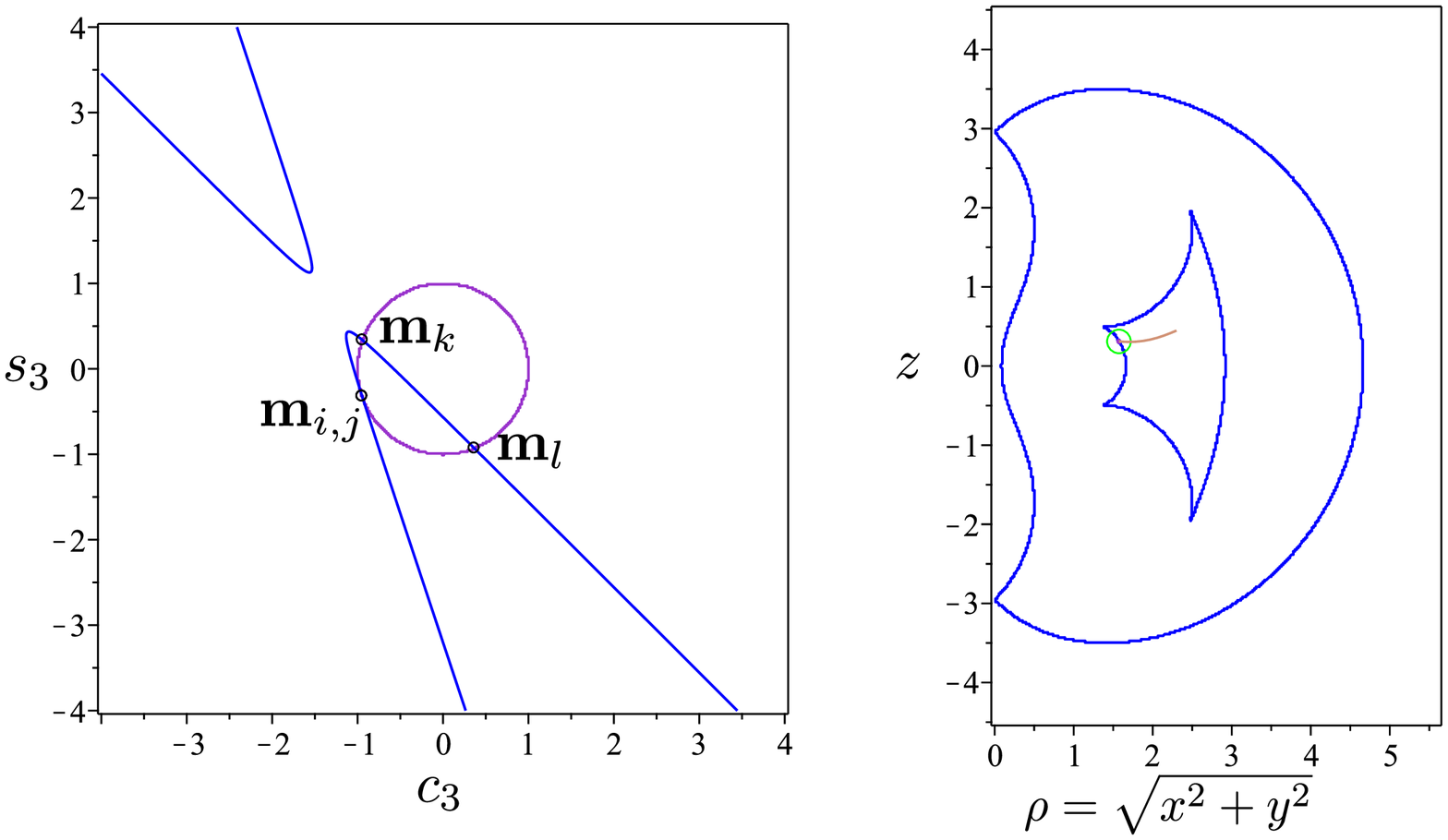}
			\caption{Tangency between adjacent points on same side of the vertex}
			\label{fig:14c}
		\end{subfigure}
		\caption{The merging of two adjacent points in a conic at a tangent point and geometrical interpretation of the components of the locus of critical values.\\Robot parameters: d = [0, 1, 0], a = [1, 2, 3/2], $\alpha$ = $\left[-\dfrac{\pi}{2}, \dfrac{\pi}{2}, 0\right]$, ($\rho, z$) = (2.5, 1) and (3, 0).}
		\label{anim:sing_traj}
	\end{figure} 
\end{proof}
As the critical values represent tangent points in $c_3s_3$-plane, a node in the locus of critical values is when we have two tangency points. A cusp occurs when three out of four intersection points merge together at a tangent point. All four solutions cannot meet together at a tangent point in a generic 3R robot \cite{pai_genericity_1992}. Also, isolated finite points of critical values cannot exist in a generic 3R robot, and thus this particular case is not considered in the context of critical values.
\begin{thm}
	In an arbitrary generic 3R robot, the inverse kinematic solutions lie always in distinct reduced aspects.
	\label{theorem:ud}
\end{thm}
\begin{proof}
	As shown in Lemma \ref{lemma:conic_orientation}, the orientation of the conic corresponding to a particular set of D-H parameters remains constant. Suppose that the inverse kinematic solutions do not belong to distinct reduced aspects.  Then, there should exist a path between two inverse kinematic solutions without intersecting a pseudosingularity or the locus of critical points. The interpretation of such a path in $c_3s_3$-plane is that two intersection points $\mathbf{m}_j$ and $\mathbf{m}_l$ switch places and neither $\mathbf{m}_j$ nor $\mathbf{m}_l$ becomes a tangent point in $c_3s_3$-plane. \\
	As the orientation of the principle axes of the conic does not change, the intersection points, $\mathbf{m}_j$ and $\mathbf{m}_l$, in $c_3s_3$-plane cannot be adjacent in a cyclic ordering since it would imply $\mathbf{m}_j$ meeting $\mathbf{m}_l$ at a tangent point to switch with $\mathbf{m}_l$. Let $\mathbf{m}_k$ lie between $\mathbf{m}_j$ and $\mathbf{m}_l$ while travelling clockwise starting from $\mathbf{m}_j$ and let $\mathbf{m}_i$ lie between $\mathbf{m}_j$ and $\mathbf{m}_l$ while travelling counterclockwise starting from $\mathbf{m}_j$ as shown in Fig. \ref{fig:12a}. As we know that the conic is not rotating, the only way for $\mathbf{m}_j$ to switch to $\mathbf{m}_l$ is to meet either $\mathbf{m}_k$ or $\mathbf{m}_i$ at a tangent point. If not, then $\mathbf{m}_l$ meets $\mathbf{m}_k$ or $\mathbf{m}_i$ at a tangent point. This \changes{contradicts} the assumption that the inverse kinematic solutions associated with $\mathbf{m}_j$ and $\mathbf{m}_l$ do not lie in distinct reduced aspects.
\end{proof}

In a cuspidal robot, at least two inverse kinematic solutions \changes{ $\mathbf{q}_1$ and $\mathbf{q}_2$ lie in an aspect}. By using Theorem \ref{theorem:ud}, we know that $\mathbf{q}_1$ and $\mathbf{q}_2$ lie in two separate reduced aspects, and thus we cross pseudosingularities during \changes{a} nonsingular change of posture. As pseudosingularities separate the reduced aspects whose image in the workspace lie in distinct non-connected regions (from Proposition \ref{propos:jwsing}), we cross the pseudosingularities at least twice in a nonsingular change of posture.

In order to discuss the path corresponding to the nonsingular change of posture in the workspace, components of the locus of critical values are discussed. The interpretation of these components in $c_3s_3$-plane allows one to draw important conclusions about the nature of the locus of critical values.

\subsection{Components of the locus of critical values}
A $n$-solution region in the workspace is always bounded by the locus of critical values which, for a generic 3R serial chain, can include cusps and/or nodes

\begin{defin}
	The components of the critical values are the connected components of the locus of the critical values, upon excluding all cusps and nodes. 
\end{defin}   

A point $\mathbf{p}$ in a region of the workspace with four preimages, corresponds to a situation where the conic intersects the unit circle at four points \changes{in} $c_3s_3$-plane (Fig. \ref{fig:12a}). Let the intersection points be $\mathbf{m}_i$, $\mathbf{m}_j$, $\mathbf{m}_k$ and $\mathbf{m}_l$. There are up to four different pairs in which the points can merge, viz. $\mathbf{m}_i\mathbf{m}_j$, $\mathbf{m}_j\mathbf{m}_k$, $\mathbf{m}_k\mathbf{m}_l$ and $\mathbf{m}_l\mathbf{m}_i$  (Fig. \ref{anim:sing_traj}). Thus, depending upon the type of robot, a 4-solution region in the workspace can be bounded by a maximum of four distinct components of singularities. A geometrical interpretation of the component of critical values is associated with the merging of a particular pair of intersection points in $c_3s_3$ - plane.

\begin{lemma}
	Let $\mathbf{q}_1$ and $\mathbf{q}_2$ be two inverse kinematic solutions in the same aspect. Considering a generic nonsingular change of posture from $\mathbf{q}_1$ to $\mathbf{q}_2$, the images of the pseudosingularities that the point $\mathbf{q}_1$ crosses to go to $\mathbf{q}_2$, belong to at least 2 different components of the critical values in the workspace. 
	\label{lemma:2sing_cross}
\end{lemma}

\begin{figure}[htbp]
	\centering
		\begin{subfigure}{0.66\textwidth}
			\centering
			\includegraphics[width=\textwidth]{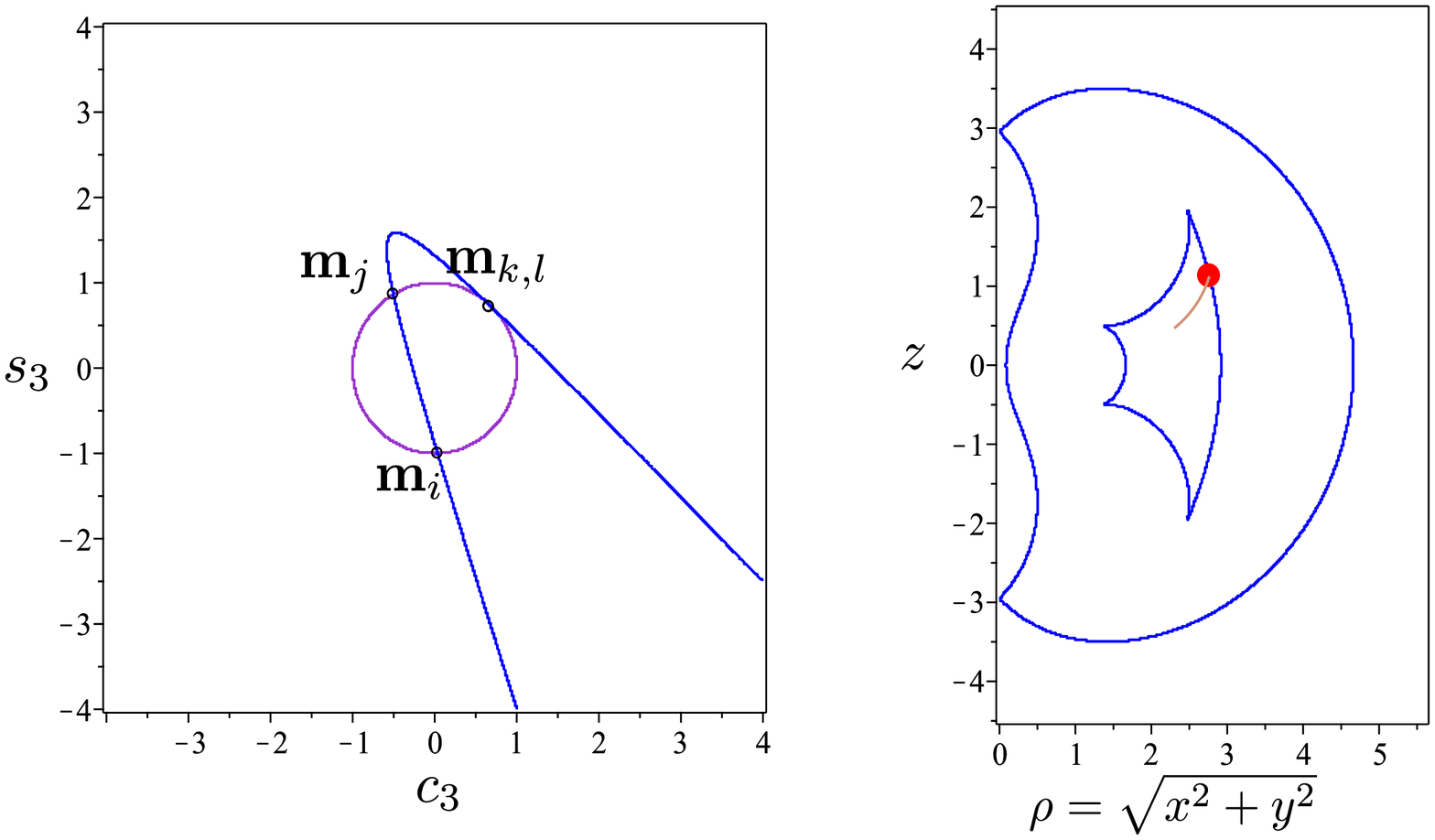}
			\caption{Component of critical value adjacent to a cusp point}
			\label{fig:15a}
		\end{subfigure}
		\\
		\begin{subfigure}{0.66\textwidth}
			\centering
			\includegraphics[width=\textwidth]{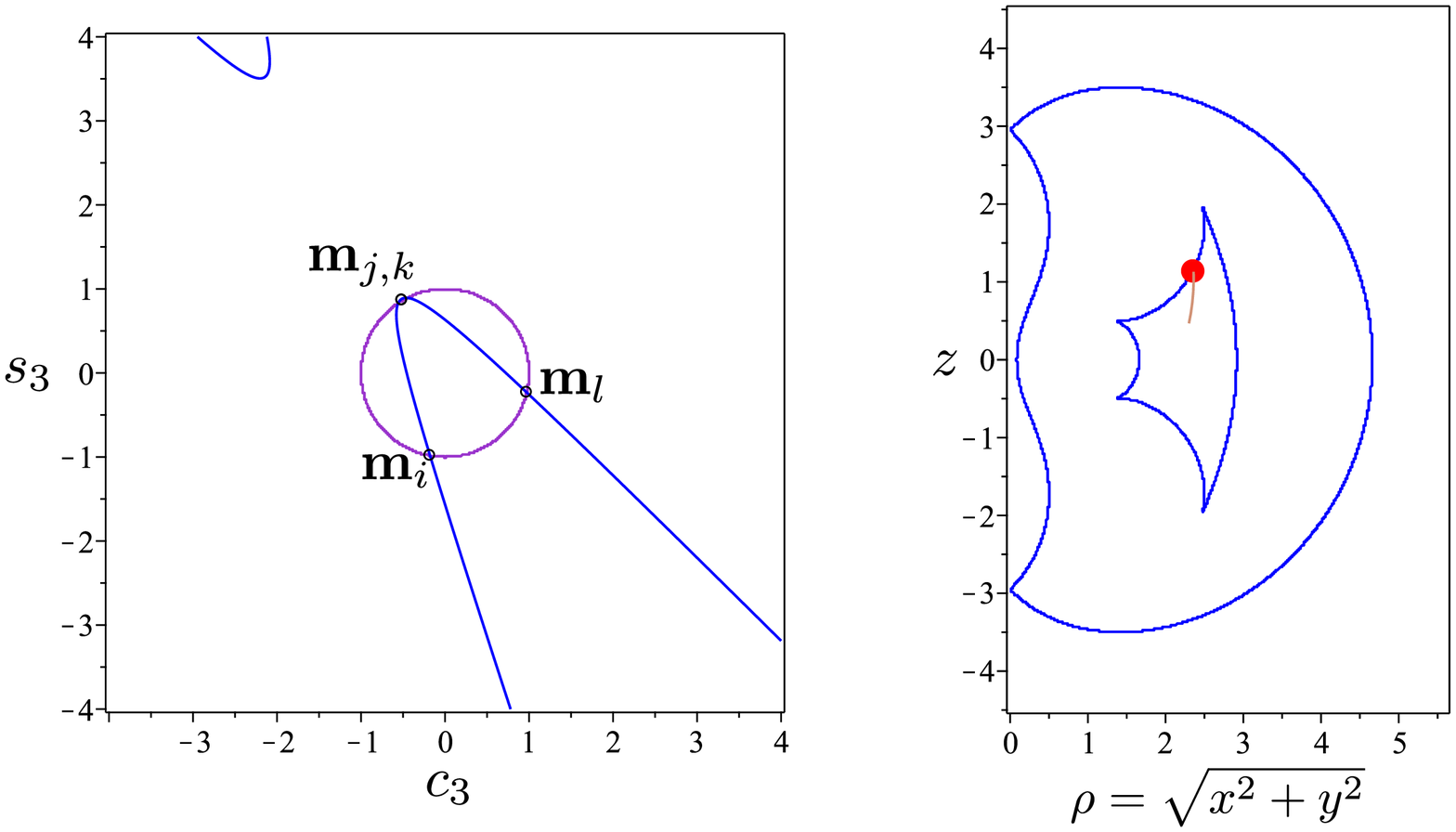}
			\caption{Component of critical value adjacent to a cusp point}
			\label{fig:15b}
		\end{subfigure}
		\\
		\begin{subfigure}{0.66\textwidth}
			\centering
			\includegraphics[width=\textwidth]{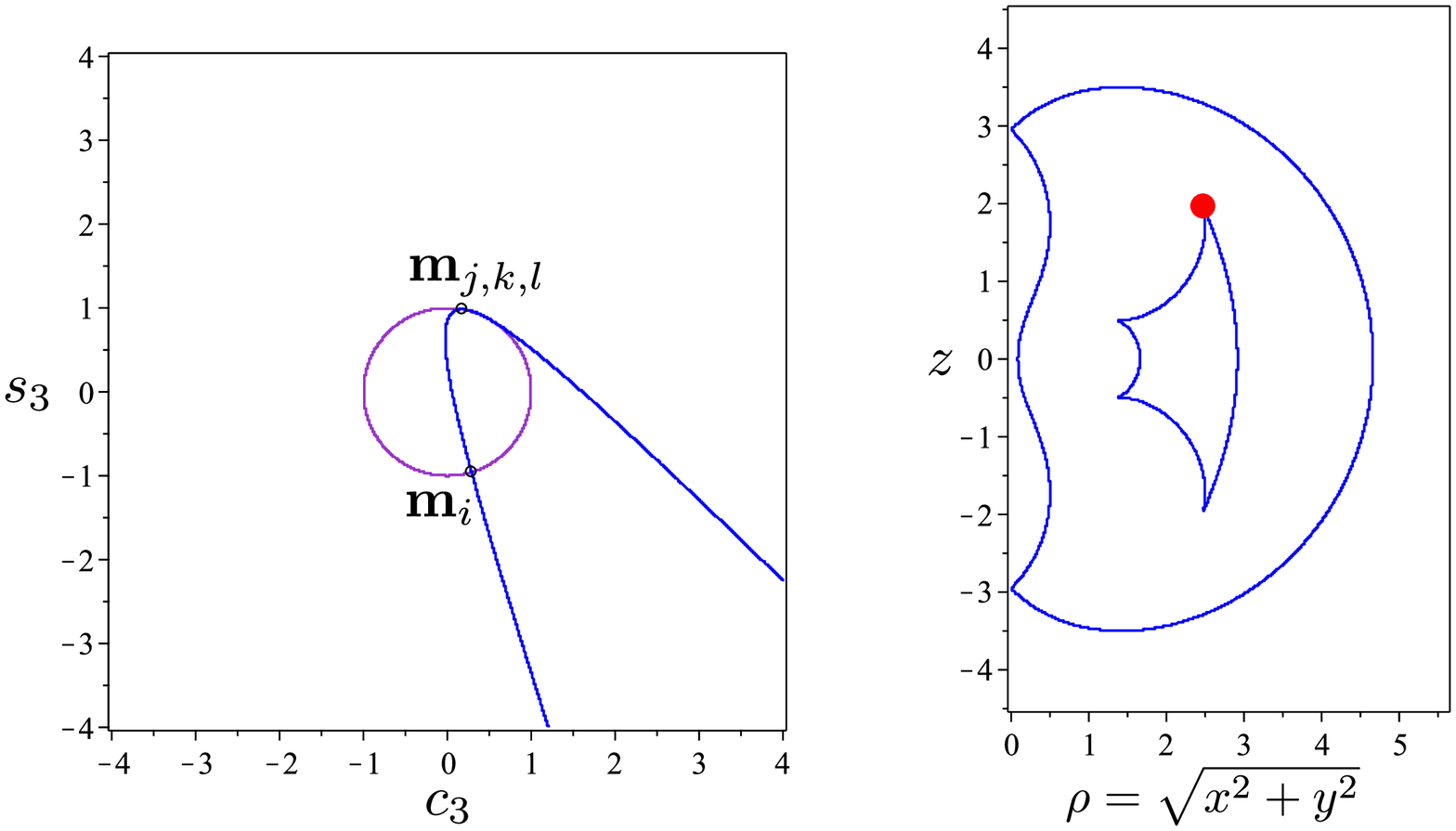}
			\caption{Cusp point}
			\label{fig:15c}
		\end{subfigure}
	\captioncomment{Geometrical interpretation of the cusp point and the adjacent components of critical values}{Robot parameters: d = [0, 1, 0], a = [1, 2, 3/2], $\alpha$ = [-$\dfrac{\pi}{2}$, $\dfrac{\pi}{2}$, 0].}
	\label{fig:cusp_ws_conic}
\end{figure}

\begin{proof}
	It is evident from the definition of pseudosingularity curve that if an IKS of a robot lies on the pseudosingularity curve, then there exists an IKS of the robot on the locus of critical points as well. An example of nonsingular change of posture is shown in Fig. \ref{fig:lemma5} where the path crosses the pseudosingularity curve twice. $\mathbf{q}_j$ crosses the pseudosingularity curve twice at $\mathbf{ps}_1$ and $\mathbf{ps}_2$ in order to switch with $\mathbf{q}_l$ in a nonsingular posture change. From Theorem \ref{theorem:ud}, we know that $\mathbf{q}_j$ and $\mathbf{q}_l$ lie in two \changes{distinct reduced aspects $\mathcal{A}$ and $\mathcal{B}$, respectively}. The reduced aspect $\mathcal{A}$ is bounded by the locus of critical points and at least by the segment of the pseudosingularity curve including $\mathbf{ps}_1$. The reduced aspect $\mathcal{B}$ is bounded by the locus of critical points and at least by the segment of the pseudosingularity curve including $\mathbf{ps}_2$. By Proposition \ref{propos:jwsing}, we assert that in generic 3R robots, pseudosingularities always separate the reduced aspects whose image in the workspace belong to regions with different number of IKS. So, we know that when $\mathbf{q}_j$ crosses $\mathbf{ps}_1$, $\mathbf{q}_l$ disappears after meeting the locus of critical points bounding the reduced aspect $\mathcal{B}$. Clearly, $\mathbf{q}_j$ crosses the pseudosingularity at $\mathbf{ps}_2$ in order to enter the reduced aspect $\mathcal{B}$. For each point in the reduced aspect $\mathcal{B}$, there should be a corresponding point in the reduced aspect, $\mathcal{A}$, as both of them map to the same bounded region in the workspace, as shown in Fig. \ref{fig:UDE}. Thus, when $\mathbf{q}_j$ is on $\mathbf{ps}_2$, there appears a point corresponding to $\mathbf{q}_l$ on the locus of critical points bounding the reduced aspect $\mathcal{A}$. 
	
	Let $\mathbf{m}_i, \mathbf{m}_j, \mathbf{m}_k, \mathbf{m}_l$ be the four intersection points in $c_3s_3$-plane corresponding to the four IKS in the joint space, $\mathbf{q}_i, \mathbf{q}_j, \mathbf{q}_k, \mathbf{q}_l$ (\changes{see} Fig. \ref{fig:lemma5}). When $\mathbf{q}_j$ crosses the pseudosingularity curve at $\mathbf{ps}_1$, $\mathbf{q}_l$ meets either $\mathbf{q}_i$ or $\mathbf{q}_k$. In $c_3s_3$-plane, \changes{$\mathbf{q}_l$} meeting $\mathbf{q}_i$ or $\mathbf{q}_k$ is similar to $\mathbf{m}_l$ meeting either $\mathbf{m}_i$ or $\mathbf{m}_k$ at the tangent point. Now, when $\mathbf{q}_j$ crosses the pseudosingularity curve at $\mathbf{ps}_2$, $\mathbf{q}_l$ enters $\mathcal{A}$. \changes{$\mathbf{q}_l$} emerging on the locus of critical point bounding $\mathcal{A}$ is similar to $\mathbf{m}_l$ merging with $\mathbf{m}_i$ or $\mathbf{m}_k$ in the initial setup. Thus, the images of the critical points bounding $\mathcal{A}$ and $\mathcal{B}$ belong to two separate components of critical values as $\mathbf{q}_l$ switches position with $\mathbf{q}_j$. This proves that the images corresponding to $\mathbf{ps}_1$ and $\mathbf{ps}_2$ lie on two distinct components of critical values in the workspace.
\end{proof}
\begin{figure}
	\centering
	\begin{subfigure}{0.9\textwidth}
		\centering
		\includegraphics[width = \textwidth]{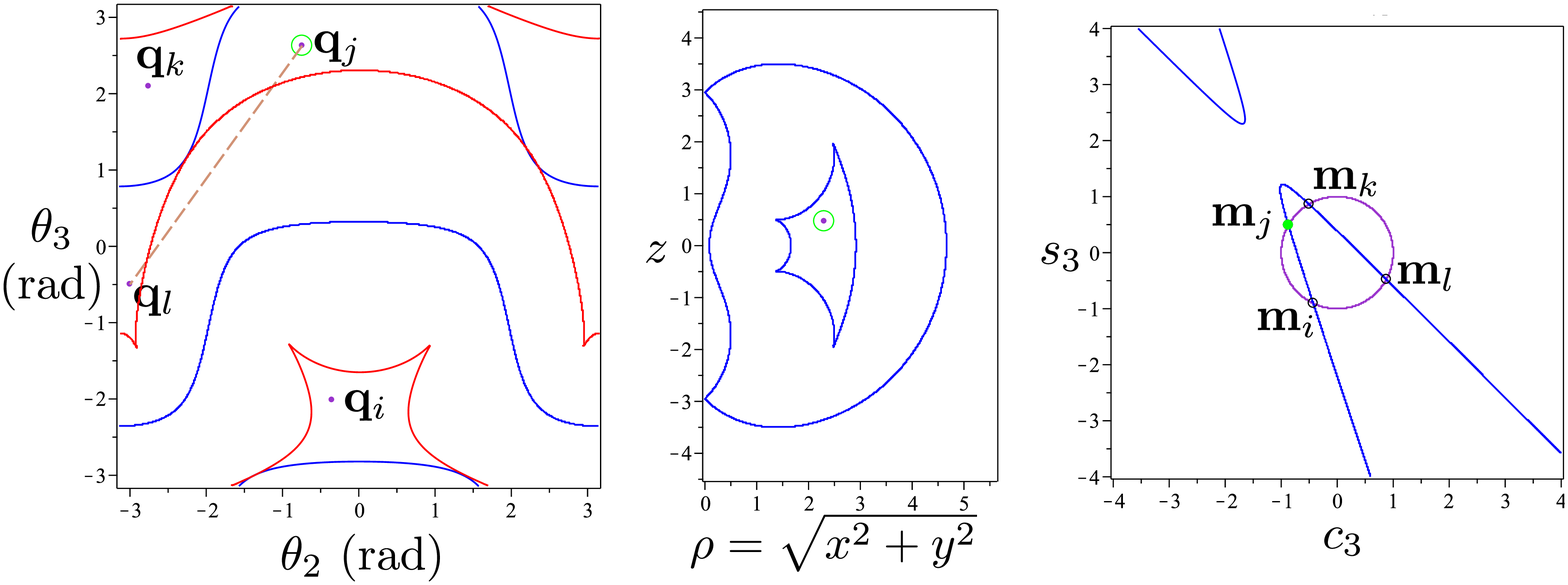}
		\caption{Phase 1: Starting from a point in workspace with 4 IKS, $\mathbf{q}_j$ and $\mathbf{q}_l$ are in same aspect.}
		\label{fig:fig_17a}
	\end{subfigure}
	\\
	\begin{subfigure}{0.9\textwidth}
		\centering
		\includegraphics[width = \textwidth]{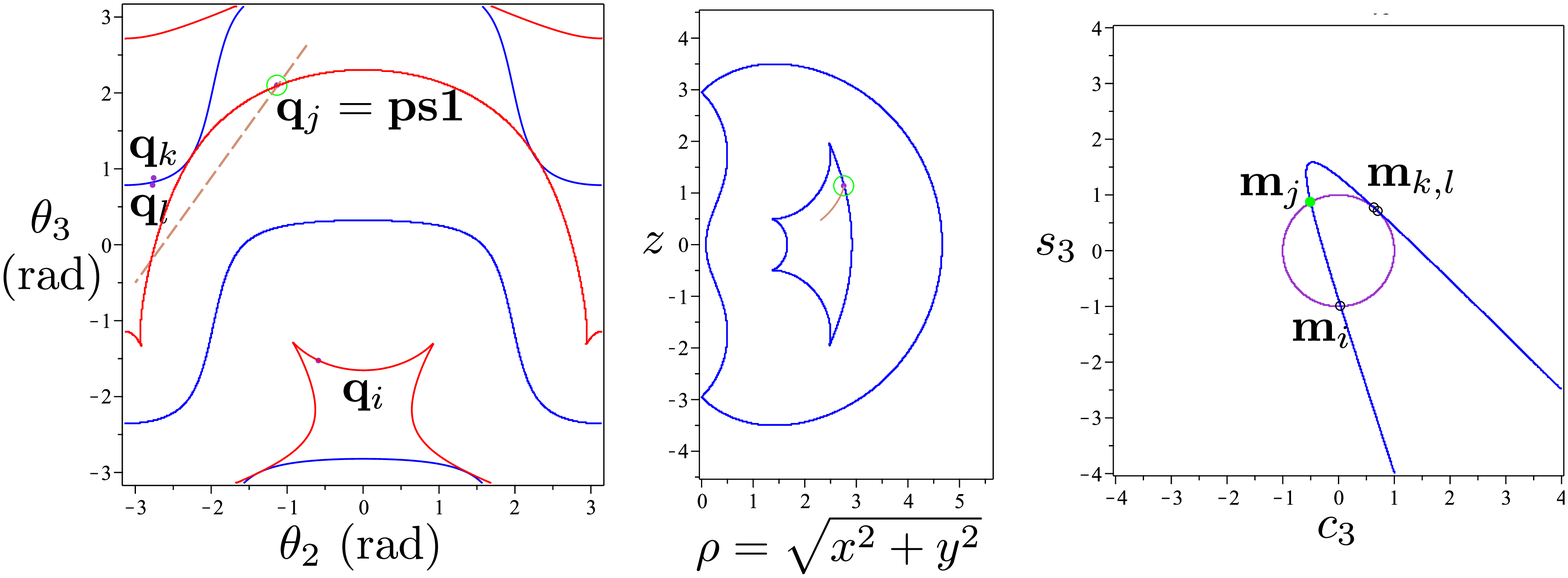}
		\caption{Phase 2: $\mathbf{q}_j$ meets $\mathbf{ps}_1$ and $\mathbf{q}_l$ meets $\mathbf{q}_k$ at singularity curve.}
		\label{fig:fig_17b}
	\end{subfigure}
	\\
	\begin{subfigure}{0.9\textwidth}
		\centering
		\includegraphics[width = \textwidth]{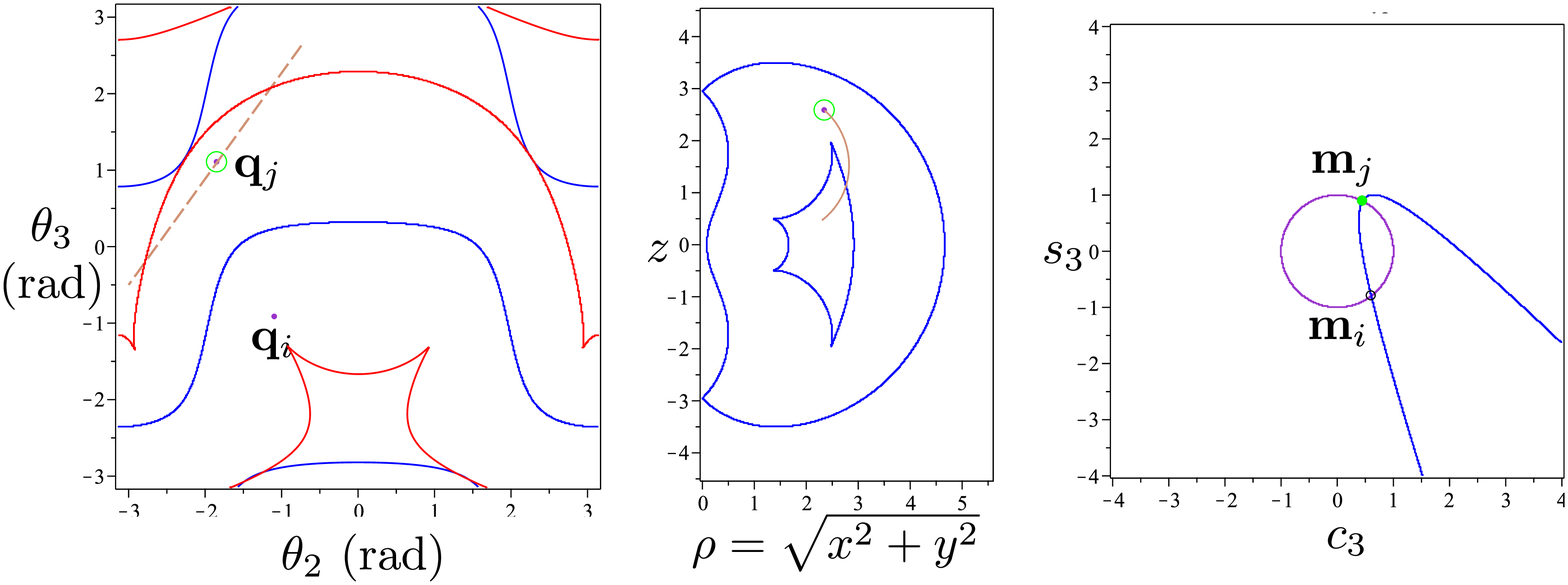}
		\caption{Phase 3: $\mathbf{q}_j$ enters another reduced aspect and $\mathbf{q}_l$ disappears.}
		\label{fig:fig_17c}
	\end{subfigure}
	\end{figure}%
    \begin{figure}[ht]\ContinuedFloat
    \centering
    \begin{subfigure}{0.9\textwidth}
		\centering
		\includegraphics[width = \textwidth]{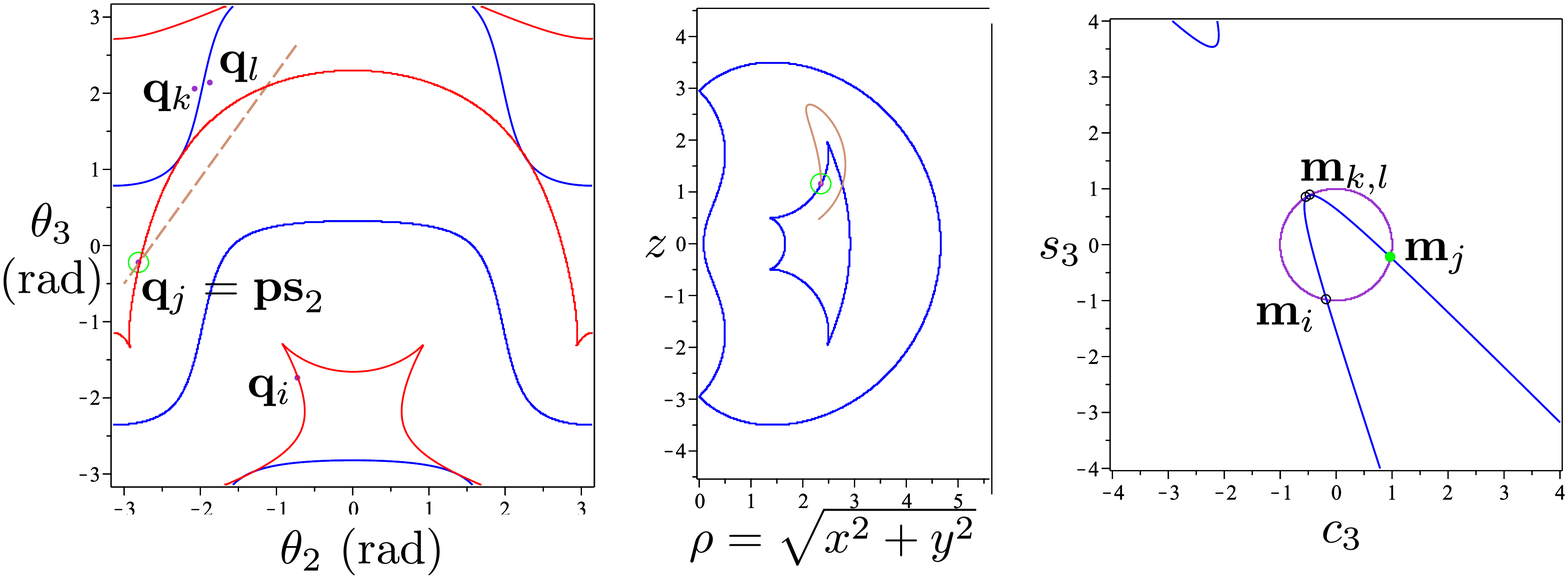}
		\caption{Phase 4: $\mathbf{q}_j$ meets $\mathbf{ps}_2$ and $\mathbf{q}_l$, $\mathbf{q}_k$ reappear, but in different reduced aspects.}
		\label{fig:fig_17d}
	\end{subfigure}
    	\\
	\begin{subfigure}{0.9\textwidth}
		\centering
		\includegraphics[width = \textwidth]{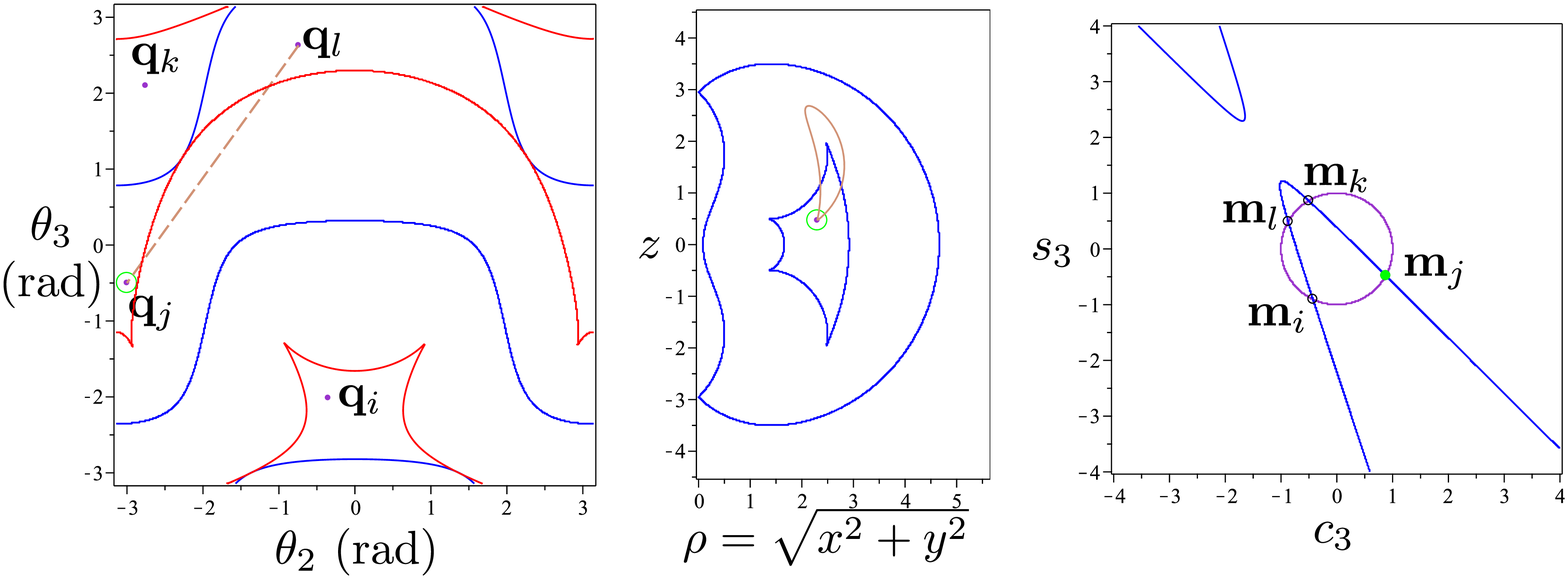}
		\caption{Phase 5: $\mathbf{q}_j$ switches with $\mathbf{q}_l$ without disappearing.}
		\label{fig:fig_17e}
	\end{subfigure}
	\caption{An example of nonsingular change of posture crossing a pseudosingularity curve at 2 points.\\Robot parameters: d = [0, 1, 0], a = [1, 2, 3/2], $\alpha$ = [$-\dfrac{\pi}{2}$, $\dfrac{\pi}{2}$, 0].\\path in the joint space ($\theta_2$, $\theta_3$): from ($-0.742, 2.628$) to ($-3, -0.5$).}
    \label{fig:lemma5}
\end{figure} 
So a nonsingular change of posture in the workspace looks like a path that exits the 4-solution region by crossing a component of critical values and \changes{re-entering the region} by crossing another component of critical values. Figure \ref{fig:nspc_wscs} shows an example of a point crossing two components of critical values in a nonsingular posture change. The path does not necessarily enter and exit \changes{by crossing} the components of critical values that form a cusp (refer to Fig. \ref{fig:2cuspJSWS}), but it is imperative to note that by proving Lemma \ref{lemma:2sing_cross}, we know that the path has to exit and enter by crossing two distinct components of critical value.
\begin{lemma}
	If points $\mathbf{m}_j$, $\mathbf{m}_l$ in $c_3s_3$-plane belong to the same aspect in the joint space and \changes{if} there exists a cusp in the workspace, then there exists another intersection point, $\mathbf{m}_k$, in the middle of (in terms of circle ordering) $\mathbf{m}_j$ and $\mathbf{m}_l$, such that $\mathbf{m}_j$,$\mathbf{m}_k$ and $\mathbf{m}_k$,$\mathbf{m}_l$ correspond to the two components of critical values that form a cusp in the workspace (see Fig. \ref{fig:cusp_ws_conic}).
\end{lemma}
\begin{proof}
	As shown in Fig. \ref{fig:tangent_case}, the interpretation of a cusp in $c_3s_3$-plane is such that 3 intersection points come together at a tangent point. Also, a cusp is a merging point of two separate components of critical value. As the components of critical values relate to merging of a particular pair of intersection points, if two components are meeting at a cusp in the workspace such that $\mathbf{m}_j, \mathbf{m}_k$ and $\mathbf{m}_l$ merge in $c_3s_3$-plane with $\mathbf{m}_k$ being in between $\mathbf{m}_j$ and $\mathbf{m}_l$, then the two components of critical values must belong to the merging of $\mathbf{m}_j\mathbf{m}_k$ and $\mathbf{m}_k\mathbf{m}_l$.
\end{proof}

\begin{lemma}
	For a nonsingular change of posture starting from a point in a bounded region $\mathcal{A}_w$ in the workspace, there exists a path in the workspace that does not meet any critical values not bounding region $\mathcal{A}_w$.
	\label{lemma:ws_traj}
\end{lemma}

\begin{proof}
    If there exists a cusp, the Lemma is automatically true~\cite{wenger_changing_1996} as encircling the cusp point is an example of a path that intersects the components of critical values bounding a single region in the workspace. We thus consider only the case in which there is no cusp in the boundary of the region $\mathcal{A}_w$. The proof of the Lemma comes from the simultaneous analysis of the nonsingular change of posture in the workspace as well as in $c_3s_3$-plane. Considering a point $\mathbf{p}$ in the 4-solution region in the workspace (\changes{see} Fig. \ref{fig:candy_case1}), let the four intersection points corresponding to this position in $c_3s_3$-plane be $\mathbf{m}_i$, $\mathbf{m}_j$, $\mathbf{m}_k$ and $\mathbf{m}_l$. Let, $\mathbf{m}_i$ and $\mathbf{m}_k$ be the solutions in the same aspects. We know from Lemma \ref{lemma:2sing_cross} that the region $\mathcal{A}_w$ is bounded by at least 2 components of critical values. As the boundary of the region $\mathcal{A}_w$ does not have a cusp, the components of critical values that bound the region $\mathcal{A}_w$ in the workspace are related to two cases: merging of $\mathbf{m}_i$, $\mathbf{m}_l$ and $\mathbf{m}_j$, $\mathbf{m}_k$ (\changes{see} Fig. \ref{fig:candy_case1}) or $\mathbf{m}_i$, $\mathbf{m}_j$ and $\mathbf{m}_k$, $\mathbf{m}_l$. Without loss of generality, we may assume the first case. As there are only two possible tangent points, $\mathcal{A}_w$ is bounded by only 2 components of critical values.
    \begin{figure}%[htbp]
		\centering
		\includegraphics[height= 0.25\textheight]{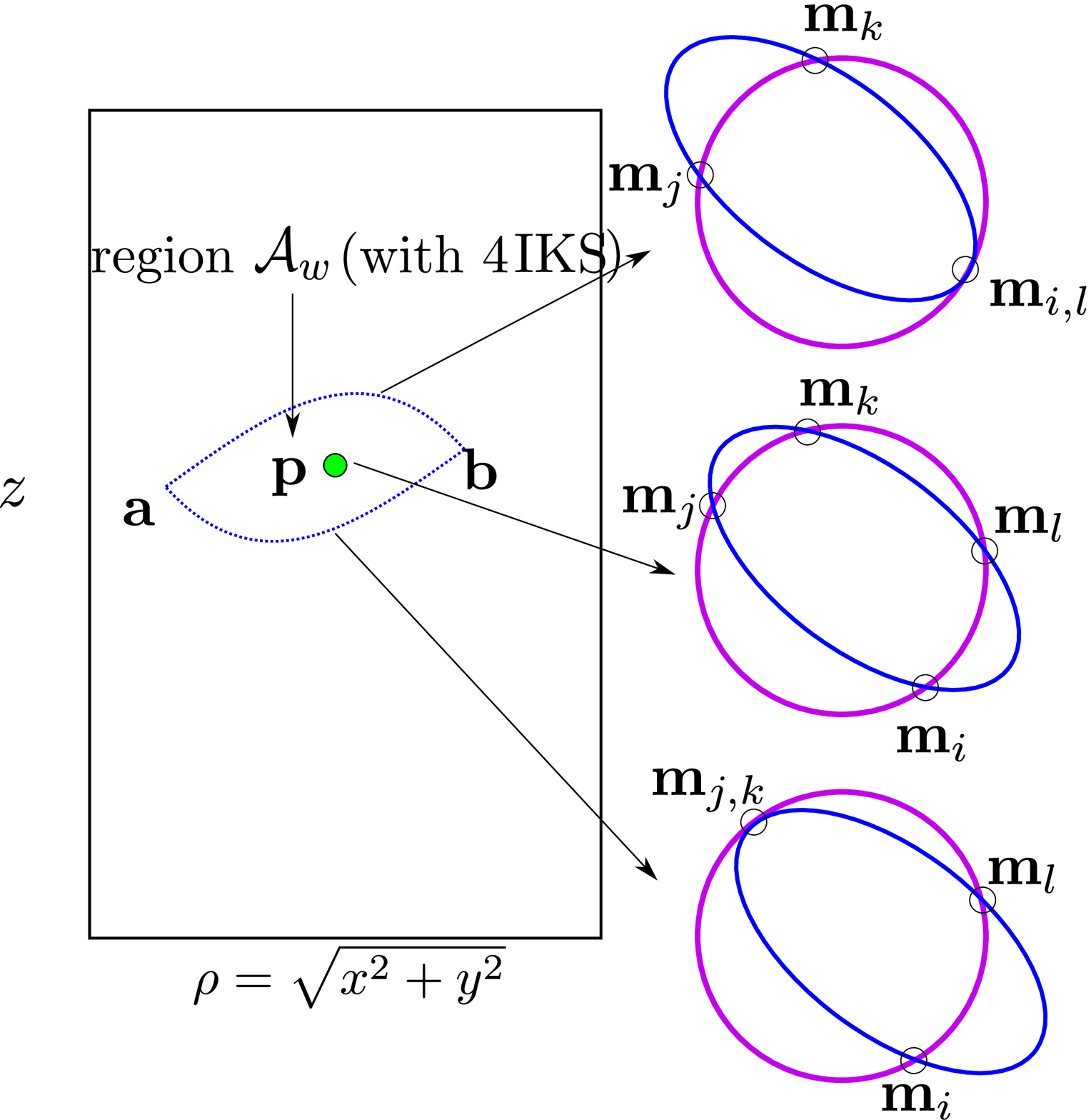}
		\caption{Region $\mathcal{A}_w$ in the workspace with 4 IKS and its geometrical interpretation. Robot parameters are imagined to illustrate a particular case. The conic can be a hyperbola or an ellipse.}
		\label{fig:candy_case1}
	\end{figure}
	\begin{figure}%[htbp]
		\centering
		\includegraphics[height = 0.2\textheight]{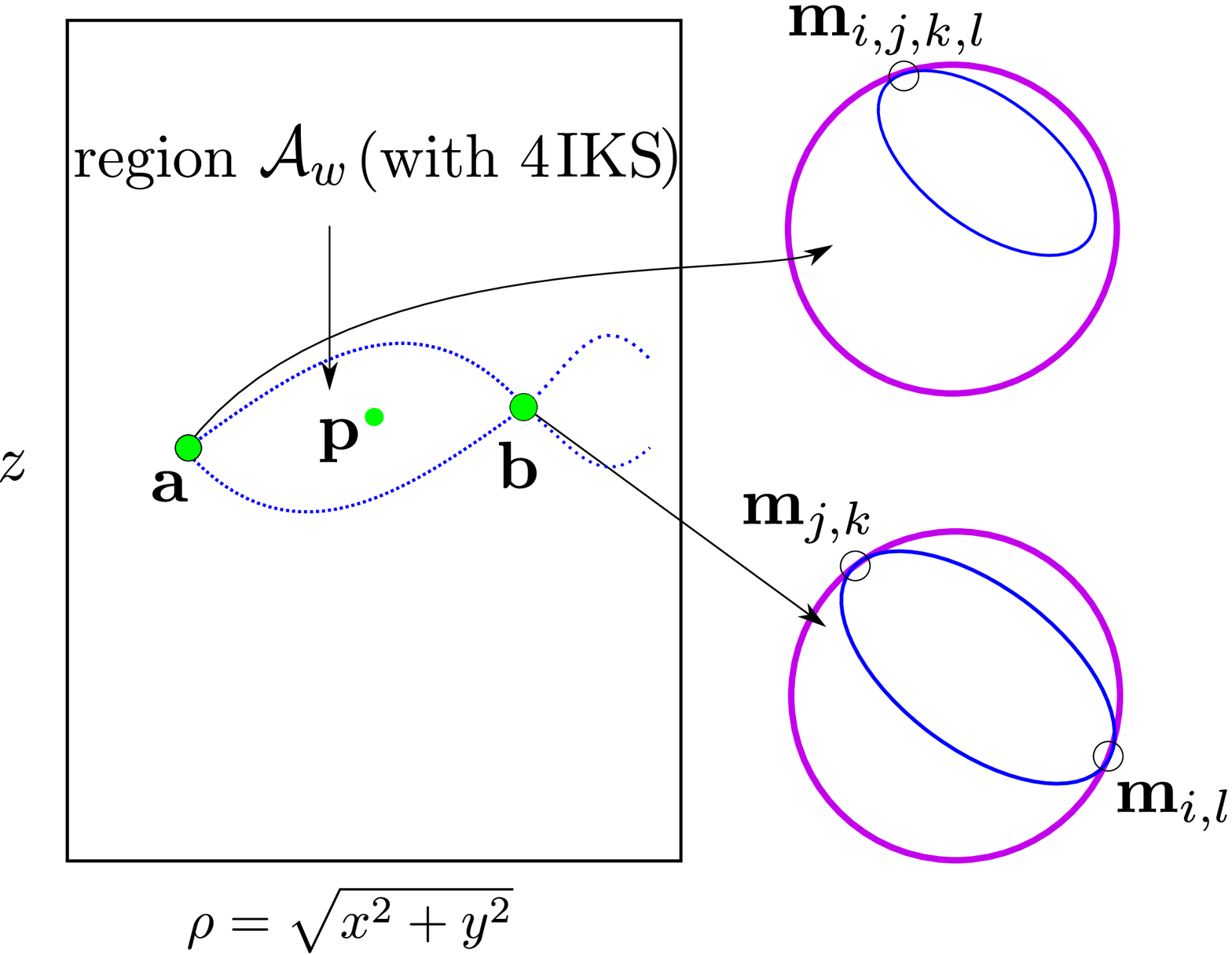}
		\caption{The intersections of components of critical values bounding $\mathcal{A}_w$ in the workspace and its geometrical interpretation. Robot parameters are imagined to illustrate a particular case. The conic can be a hyperbola or an ellipse.}
		\label{fig:candy_case2}
	\end{figure}
	
    As region $\mathcal{A}_w$ is bounded and there exists no cusps, the two components of critical values bounding the region $\mathcal{A}_w$ must intersect. In the $c_3s_3$-plane, points ($\mathbf{m}_i$,$\mathbf{m}_l$) and ($\mathbf{m}_j$,$\mathbf{m}_k$) either meet simultaneously at two distinct points (bitangent case) or they meet at a single point of tangency between the conic and the circle as illustrated in Fig. \ref{fig:candy_case2}. Since we are considering generic robots, we cannot have four equal IKS, and we can immediately conclude that the intersection of the components of critical values corresponds to the bitangent case. 
	
    Let $\mathcal{B}_w$ be an arbitrary 4 solution region in the workspace that is not $\mathcal{A}_w$. Proceeding by contradiction, it is sufficient to show that a path crossing two distinct components of the critical values bounding region $\mathcal{B}_w$ in order to enter and exit $\mathcal{B}_w$, does not correspond to a nonsingular change of posture. An example of such a workspace is illustrated in Fig. \ref{fig:bounding_curves} and the interpretation of a closed loop path in the workspace is given in Fig. \ref{fig:invalid_path}, which shows that such a path cannot define a nonsingular change of posture.
	\begin{figure}[htbp]
		\centering
		\includegraphics[width= 0.4\textwidth]{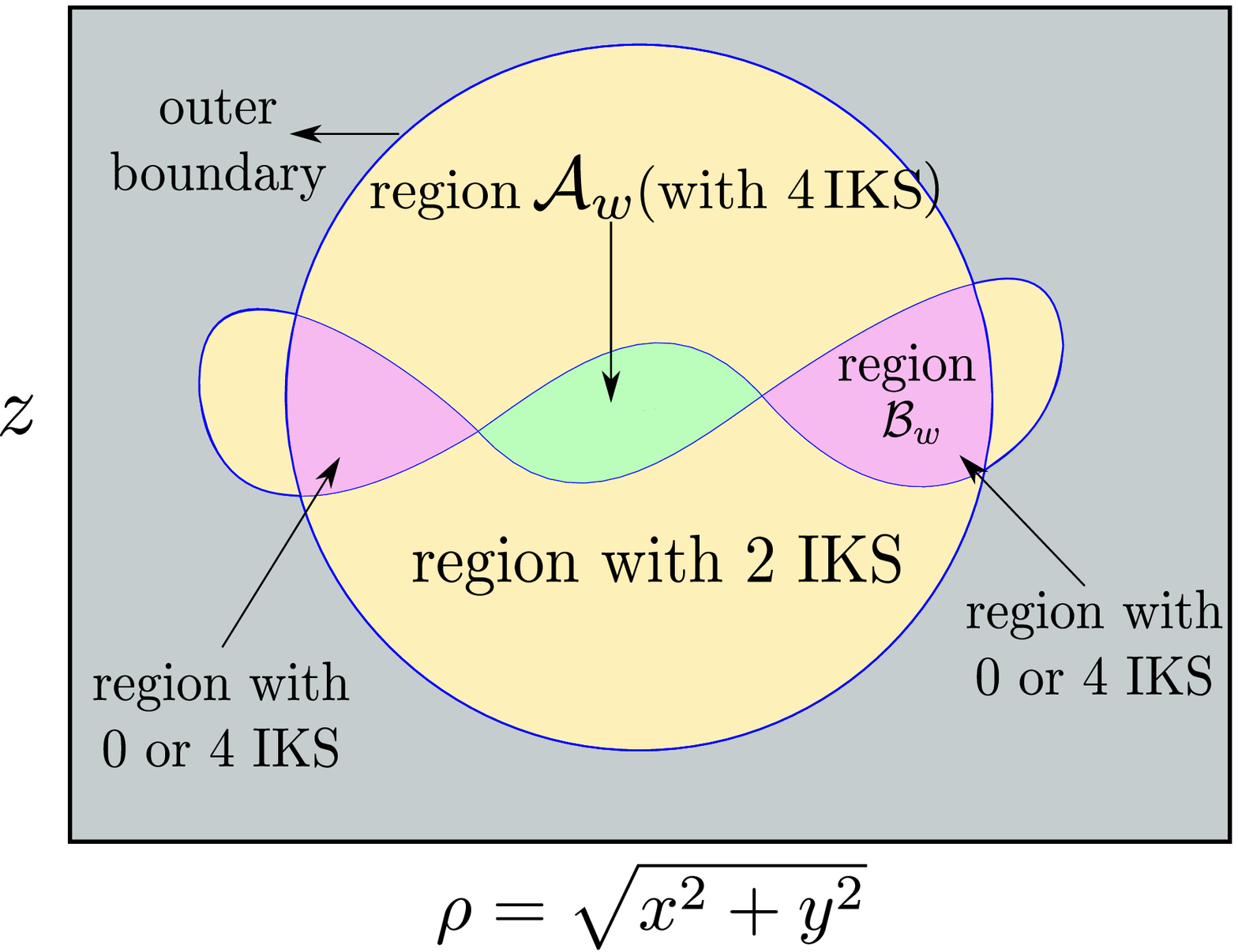}
		\caption{An example of the shape of the workspace, where a closed loop path starting from a point in $\mathcal{A}_w$ must cross two distinct components of critical values bounding region $\mathcal{B}_w$}.
		\label{fig:bounding_curves}
	\end{figure}
	From Lemma \ref{lemma:2sing_cross}, we already know that the path in workspace corresponding to the nonsingular change of posture crosses two different components of critical values of region $\mathcal{A}_w$. If the path exits $\mathcal{A}_w$ by crossing the component corresponding to the merging of $\mathbf{m}_i$ and $\mathbf{m}_l$, then it must cross the component belonging to the merging of $\mathbf{m}_j$ and $\mathbf{m}_k$ while entering the same region. Now, if the case shown in Fig. \ref{fig:bounding_curves} exists, then in order to enter $\mathcal{A}_w$, we will have to cross another 4-solution region, $\mathcal{B}_w$, bounded by at least 2 different components of critical value. As we are considering only the case without cusps in the workspace, $\mathcal{B}_w$ is also bounded by two components of critical values having no point in common, i.e. if one component corresponds to the merging of \changes{$\mathbf{m}_i$ and $\mathbf{m}_l$,} then the other component must correspond to the merging of $\mathbf{m}_j$ \changes{and} $\mathbf{m}_k$. While crossing $\mathcal{B}_w$, one needs to cross both components. This means that if one tracks the intersection point $\mathbf{m}_i$ from its initial position, then this point will have been a tangent point (in $c_3s_3$-plane) while crossing either of the two components of $\mathcal{B}_w$. This leads to a contradiction, as we have assumed that one of the points we are tracking will not be tangent to the unit circle to qualify as a valid nonsingular change of posture. Figure~\ref{fig:invalid_path} illustrates a path where two singularities of another 4-solution region without any cusp are crossed. We start from an initial point $\mathbf{p}$ in the 4-solution region, $\mathcal{A}_w$, of the workspace that corresponds to Posture 1 in the $c_3s_3$-plane (refer to Fig. \ref{fig:invalid_path}). We assume that the robot corresponding to this case is cuspidal and the IKS corresponding to $\mathbf{m}_j$ and $\mathbf{m}_l$ lie in the same aspect. In Posture 2, we cross the component of critical values that belongs to the merging of $\mathbf{m}_i$ and $\mathbf{m}_l$. This already suggests that for a valid nonsingular change of posture, $\mathbf{m}_j$ should switch places with $\mathbf{m}_l$ without being a tangent point in the $c_3s_3$-plane. The path going from Posture 3 to Posture 5 is the entry into another 4-solution region, $\mathcal{B}_w$. As we have entered region $\mathcal{B}_w$ by crossing the component of critical values corresponding to the merging of $\mathbf{m}_i$ and $\mathbf{m}_l$, we need to exit the region by crossing the component of critical values corresponding to the merging of $\mathbf{m}_j$ and, $\mathbf{m}_k$ as shown in Posture 6. This proves that such a path is an invalid nonsingular change of posture, as we encountered a singular configuration while exiting $\mathcal{B}_w$.
	\begin{figure}[htbp]
		\centering
		\includegraphics[width= 0.85\textwidth]{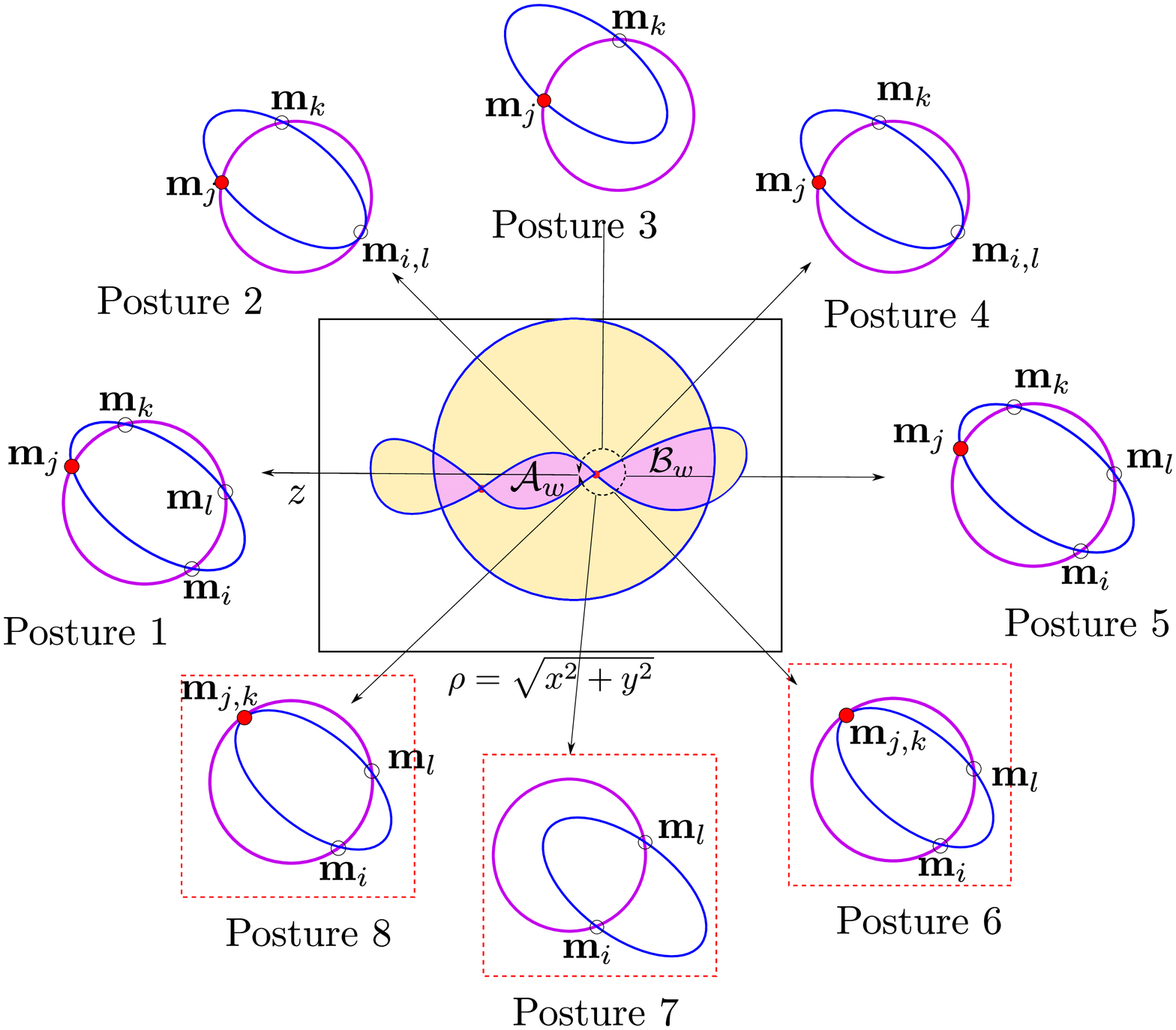}
		\caption{The closed loop path in the workspace, where the path crosses another 4-solution region and its corresponding interpretation in $c_3s_3$-plane. Robot parameters are imagined to illustrate a particular case to show that such a path does not correspond to a nonsingular change of posture. The postures in red boxes correspond to the steps in the shown path where the definition of nonsingular posture change is violated}.
		\label{fig:invalid_path}
	\end{figure}
\end{proof}

\subsection{The ``candy'' case}
\label{section:candy_case}
\begin{proof}[Proof of Proposition \ref{propos:second}]
By Lemma \ref{lemma:ws_traj}, the curves corresponding to the critical values are enclosed by an outer boundary of the workspace and would have a shape as illustrated in Fig. \ref{fig:candy_case3}. We shall refer to this shape as the ``candy'' case. We arrive at this case by starting with a point, $\mathbf{p}$, in the workspace with four preimages. Let the points of intersection in $c_3s_3$-plane corresponding to $\mathbf{p}$ be $\mathbf{m}_i$, $\mathbf{m}_j$, $\mathbf{m}_k$ and $\mathbf{m}_l$. The region, $\mathcal{A}_w$, in which $\mathbf{p}$ exists must be bounded by the locus of critical values. As we \changes{assume} that there are no cusps, the region will have to be bounded by only two components corresponding to the merging of ($\mathbf{m}_i$,$\mathbf{m}_j$ and $\mathbf{m}_k$,$\mathbf{m}_l$) or ($\mathbf{m}_i$,$\mathbf{m}_l$ and $\mathbf{m}_j$,$\mathbf{m}_k$) as shown in Fig. \ref{fig:candy_case1}. These two components of the critical values intersect at two points, $\mathbf{a}$ and $\mathbf{b}$. The geometrical interpretation of the intersection of the components of critical values is that both $\mathbf{m}_i$,$\mathbf{m}_l$ and $\mathbf{m}_j$,$\mathbf{m}_k$ merge at tangent points simultaneously. This can happen in two cases, either all four intersection points merge together or $\mathbf{m}_i$,$\mathbf{m}_l$ and $\mathbf{m}_j$,$\mathbf{m}_k$ meet together at separate tangent points in $c_3s_3$-plane, forming a node point in the workspace as shown in Fig. \ref{fig:candy_case2}. \changes{Since four points merging} at a single tangent point \changes{corresponds} to a nongeneric case, we will consider only the case of $\mathbf{m}_i$,$\mathbf{m}_l$ and $\mathbf{m}_j$,$\mathbf{m}_k$ meeting together at separate tangent points. As at the ends of our candy shape, only two segments meet without forming a node (see $\mathbf{m}$ and $\mathbf{n}$ in Fig. \ref{fig:candy_case3}), it corresponds to a case of four solutions merging at a common point. This is a contradiction to the assumption of a generic 3R robot because points with multiplicity four correspond to a nongeneric 3R robot \cite{pai_genericity_1992}.
\begin{figure}[htbp]
	\centering
	\includegraphics[width= 0.4\textwidth]{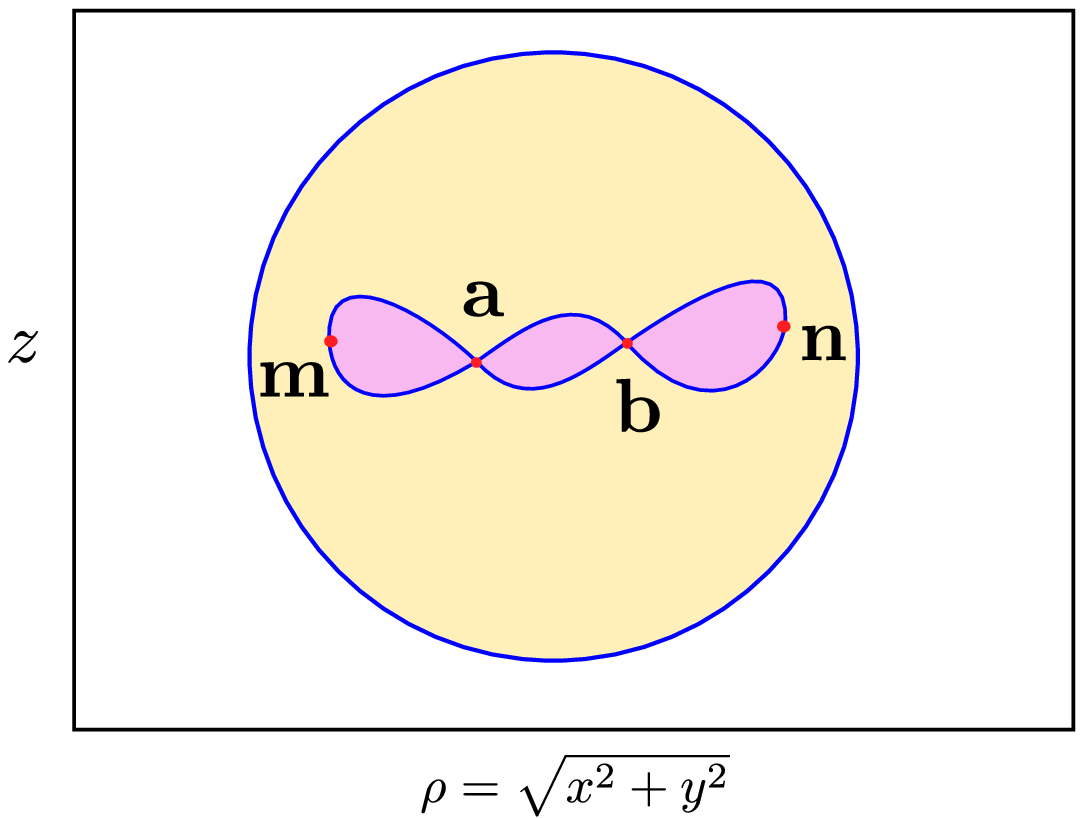}
	\caption{The ``candy'' case.}
	\label{fig:candy_case3}
\end{figure}
\end{proof}
By using \changes{Theorem} \ref{propos:first} and  Proposition \ref{propos:second}, a necessary and sufficient condition can be derived for a generic 3R cuspidal robot. Formally, the corollary is stated as:
\begin{thm}
	\changes{The} existence of a cusp point in the workspace is a necessary and sufficient condition for \changes{a generic robot} to be cuspidal.
\end{thm}

\begin{figure}
    \centering
    \begin{subfigure}{0.95\textwidth}
        \includegraphics[width=\textwidth]{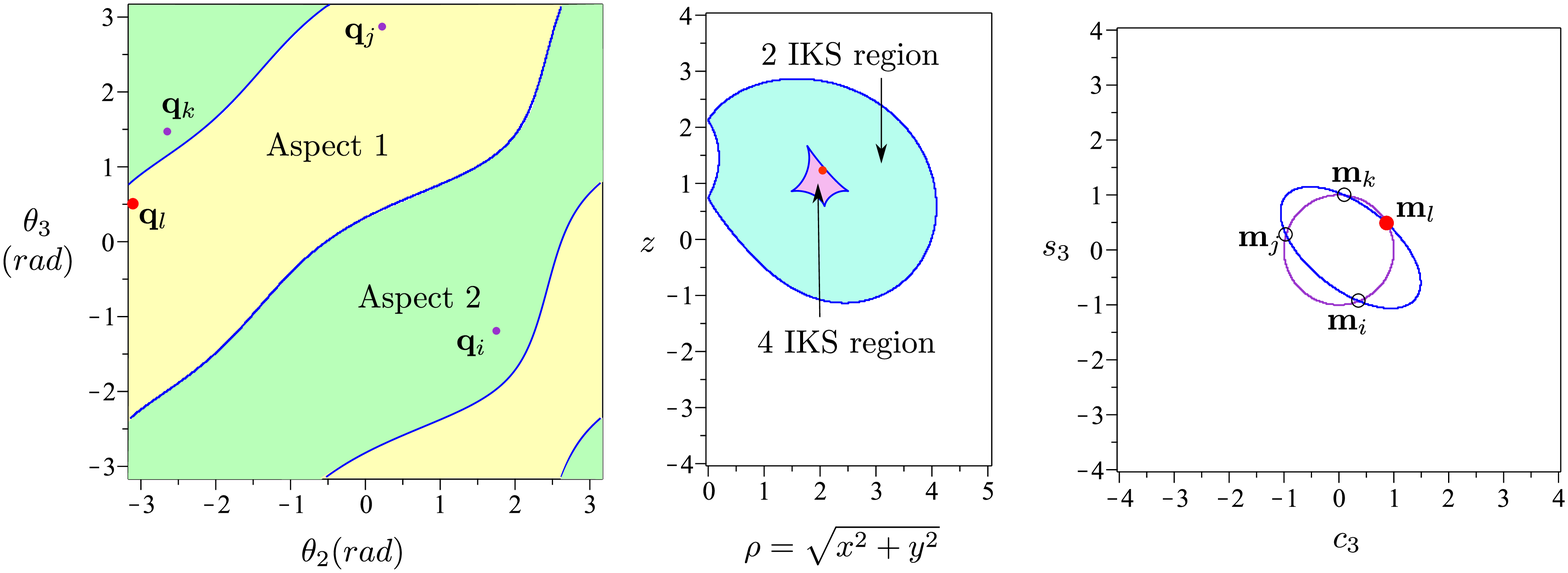}
        \caption{An example of the non-orthogonal and cuspidal case. \\ Robot parameters: Robot parameters: d = [0, 1, 0], a = [1, 2, 1], $\alpha$ = [-$\dfrac{\pi}{6}$, $\dfrac{\pi}{2}$, 0].}
        \label{fig:noc}
    \end{subfigure}
    \\
    \begin{subfigure}{0.95\textwidth}
        \includegraphics[width=\textwidth]{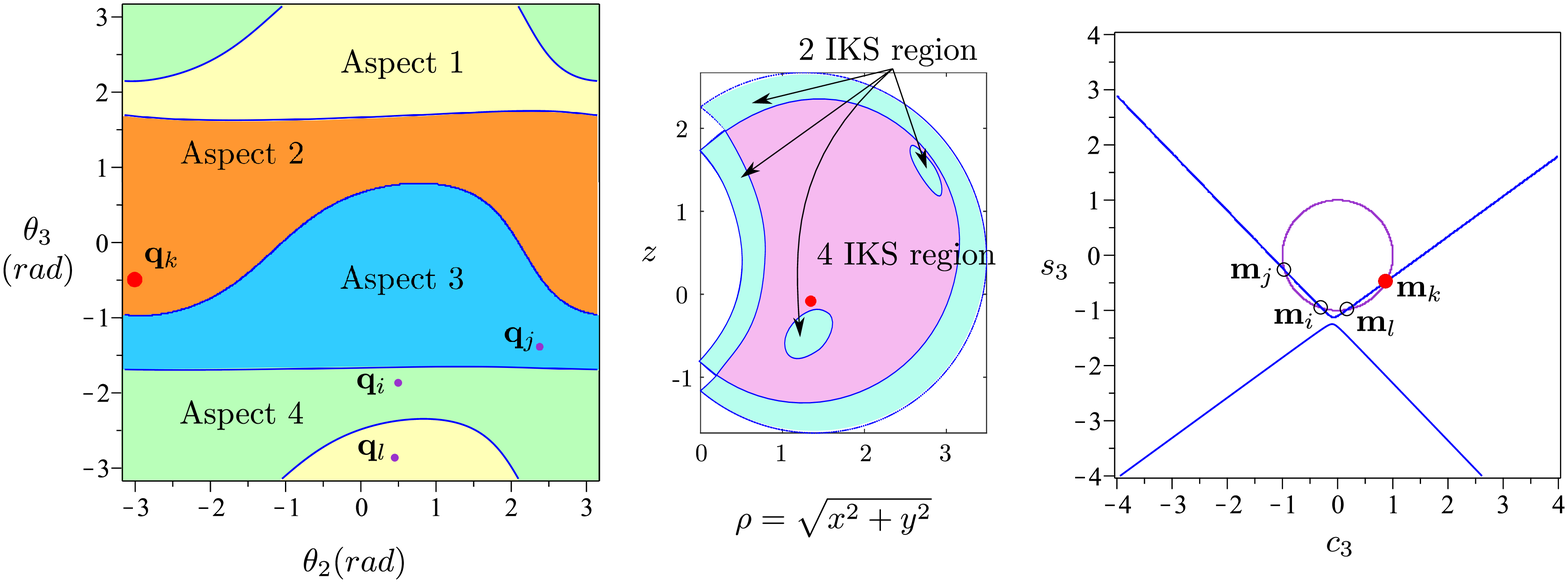}
        \caption{An example of the non-orthogonal and non-cuspidal case. \\ Robot parameters: Robot parameters: d = [0, 1, 0], a = [1, 0.2, 2], $\alpha$ = [-$\dfrac{\pi}{3}$, 1.745, 0].}
        \label{fig:nonc}
    \end{subfigure}
    \caption{The aspects in the joint space, regions in the workspace and corresponding conics in the $c_3s_3$ - plane for a cuspidal and non-cuspidal non-orthogonal robot.}
    \label{fig:no_canc}
\end{figure}

Figure \ref{fig:no_canc} illustrates an example of non-orthogonal cuspidal and non-cuspidal robots in joint space, workspace and  $c_3s_3$-plane.

 %%%%%%%%%%% section: conclusions %%%%%%%%%%%
 \section{Conclusions}
\label{section:conclusions}  
    In the presented work, we have revisited the geometrical interpretation of the inverse kinematic model of 3R serial robots. The geometrical interpretation \changes{provides a} better understanding of the critical values in the workspace, the nodes and the cusps. The paper \changes{has also presented} important observations regarding the shape and the orientation of the conic represented in $c_3s_3$-plane, corresponding to a specific set of DH parameters. By analyzing the nonsingular posture change in joint space, workspace and $c_3s_3$-plane, the authors have extended the proof \changes{of} existence of reduced aspects in the joint space of generic 3R robots. As a main contribution of the work, we have put forth the proof \changes{of} the necessary condition of existence of a cusp in the workspace for a generic 3R serial robot to be cuspidal. In combination with an existing sufficient condition, we have presented formally the necessary and sufficient condition for a generic 3R robot to be cuspidal. The presented work is expected to be of great importance for the designer in extending the classification of generic 3R serial robots based on cuspidality. When designing a new robot, knowing whether a set of given D-H parameters \changes{defines} a cuspidal robot or not is of high interest. The proposed necessary and sufficient condition based on the existence of a cusp point in the workspace can be used by the designer to verify if the robot is cuspidal or not. As the presented work is applicable to positional robots, the results obtained in the paper can be directly extended to 6R wrist-partitioned robots with wrist at the end and at the beginning. This is due to the \changes{fact} that the singularities related to the orientation are decoupled from the positional singularities in these robots. This allows a designer to analyze the positional singularities in two dimensions only. As a part of future work, the geometrical interpretation of \changes{the} inverse kinematic model of generic 6R serial robots will be studied to have similar conclusions on the cuspidal nature of generic 6R serial robots. 
    %% future works ?
\appendix

\section{Singularity in the joint space: the locus of critical points}
The singularity for a 3R serial robot in the joint space, the locus of critical points, can be derived from the relation:
\begin{equation}
    det(\mathbf{J}) = 0
    \label{eq:det_J}
\end{equation}
where, $\mathbf{J}$, is the Jacobian matrix given in \eqref{eq:jac_def}. Let, $^i\mathbf{T}_j$ be the transformation matrix of frame $j$ with respect to frame $i$. $\mathbf{e}$ be the position vector of the end-effector with respect to frame $0$. 
\begin{align}
    \mathbf{e} &= ^0\mathbf{T}_1\,^1\mathbf{T}_2\,^2\mathbf{T}_3\,\begin{bmatrix} 0 \\ 0 \\ 0 \\ 1 \end{bmatrix}
\end{align}
The Jacobian of $\mathbf{e}$ is the Jacobian matrix for the 3R serial robot.
\begin{align}
    \mathbf{J} &= \begin{bmatrix}\dfrac{\partial \mathbf{e}}{\partial \theta_1} &
    \dfrac{\partial \mathbf{e}}{\partial \theta_2} & \dfrac{\partial \mathbf{e}}{\partial \theta_3}
    \end{bmatrix}
\end{align}
The expression for \eqref{eq:det_J} is:

\begin{align*}
    det(\mathbf{J}) = \, &(((-c_3\,(a_3\,d_2\,s_2\,s_3 + a_1\,d_3)\,ca_2 + a_2\,c_2\,c_3\,d_2 + (-c_2\,d_2\,s_3^2 + c_2\,d_2)\,a_3 - a_2\,d_3\,s_2\\
    & \,s_3)\,sa_2 + c_3\,(a_1\,a_3\,s_3 - d_2\,d_3\,s_2)\,ca_2^2 + a_2\,a_3\,s_2\,s_3^2\,ca_2 + (-s_3\,(a_2\,c_2 + a_1)\,a_3 + \\
    & d_2\,d_3\,s_2)\,c_3 - a_2\,s_3\,(a_2\,c_2 + a_1))\,sa_1 + a_1\,((ca_2\,a_3\,c_2\,c_3\,s_3 + s_2\,(a_2\,c_3 +\\ 
    & (-s_3^2 + 1)\,a_3))\,sa_2 + c_2\,c_3\,d_3\,(ca_2 - 1)\,(ca_2 + 1))\,ca_1)\,a_3
\end{align*}

Here, $c_i,\, s_i,\, ca_i,\, sa_i$ correspond to $\cos\theta_i,\, \sin\theta_i,\, \cos\alpha_i,\, \sin\alpha_i$ respectively and $\alpha_i$, $d_i$ and $a_i$ are the classical D-H parameters (see fig. \ref{fig:dhpara}).

%\section{\changes{Singularity in the workspace: the locus of critical values}}

 %%%%%%%%%%% section: conclusions %%%%%%%%%%%
 \section*{Acknowledgement}
The authors are supported by the joint French and Austrian ECARP project: 
ANR-19-CE48-0015, FWF I4452-N.

\bibliographystyle{elsarticle-num}
\bibliography{references}
\end{document}